\documentclass[letterpaper]{article} 
\usepackage{stackengine}
\usepackage[draft]{aaai2026}  
\usepackage{times}  
\usepackage{helvet}  
\usepackage{courier}  
\usepackage[hyphens]{url}  
\usepackage{graphicx} 
\usepackage{natbib}  
\usepackage{caption} 
\usepackage[table]{xcolor}
\usepackage{algorithm}
\usepackage[noend]{algpseudocode}

\usepackage{newfloat}
\usepackage{listings}
\DeclareCaptionStyle{ruled}{labelfont=normalfont,labelsep=colon,strut=off} 
\floatstyle{ruled}
\newfloat{listing}{tb}{lst}{}
\floatname{listing}{Listing}
\newcommand{\mystack}[2]{\shortstack{#1\\#2}}

\title{Lessons from Training Grounded LLMs with Verifiable Rewards}
\author{
Shang Hong Sim\textsuperscript{\rm 1\rm *},
Tej Deep Pala\textsuperscript{\rm 1\rm *},
Vernon Toh\textsuperscript{\rm 1\rm *},
Hai Leong Chieu\textsuperscript{\rm 2},
Amir Zadeh\textsuperscript{\rm 3},
Chuan Li\textsuperscript{\rm 3},
Navonil Majumder\textsuperscript{\rm 1},
Soujanya Poria\textsuperscript{\rm 1}
}
\affiliations {
\textsuperscript{\rm 1}Singapore University of Technology and Design, \textsuperscript{\rm 2}DSO National Laboratories, \textsuperscript{\rm 3}Lambda Labs\\
\textsuperscript{*}Equal contribution\\
}

\usepackage{bibentry}

\usepackage{microtype}      
\usepackage{graphicx}
\usepackage{subfigure}
\usepackage{booktabs}       
\usepackage{multirow}
\usepackage[most,skins,theorems]{tcolorbox}
\usepackage{tabularx}
\usepackage{tikz}
\usetikzlibrary{tikzmark}
\usepackage{enumitem}
\usepackage{arydshln}
\usepackage{soul}

\usepackage{amsmath}
\usepackage{amssymb}
\usepackage{mathtools}
\usepackage{amsthm}

\usepackage{url}            
\usepackage{cleveref}

\usepackage[utf8]{inputenc} 
\usepackage[T1]{fontenc}    
\usepackage{amsfonts}       
\usepackage{nicefrac}       
\usepackage{pifont}

\usepackage{pgfplots}
\usepackage{subcaption}
\usepackage{xcolor}
\pgfplotsset{compat=1.18}
\definecolor{asqa}{HTML}{1f77b4}
\definecolor{qampari}{HTML}{ff7f0e}
\definecolor{eli5}{HTML}{2ca02c}

\usepackage[textsize=tiny]{todonotes}

\definecolor{myred}{rgb}{0.7, 0.3, 0.0}
\definecolor{myblue}{HTML}{054488}
\definecolor{mygreen}{HTML}{056b34}

\makeatletter
\newcommand*\myfontsize{%
  \@setfontsize\myfontsize{7}{8}%
}
\makeatother

\tcbset{
  aibox/.style={
    width=\linewidth,
    top=7pt,
    bottom=2pt,
    colback=blue!6!white,
    colframe=black,
    colbacktitle=black,
    enhanced,
    center,
    attach boxed title to top left={yshift=-0.1in,xshift=0.15in},
    boxed title style={boxrule=0pt,colframe=white,},
  }
}
\newtcolorbox{AIbox}[2][]{aibox,title=#2,#1}

\newcommand{\cdashlinelr}[1]{%
  \noalign{\vskip 2mm}  
  \cdashline{#1}        
  \noalign{\vskip 2mm}  
}
\makeatother

\newcommand{\hlinelr}[1]{%
  \noalign{\vskip 2mm}  
  \hline    
  \noalign{\vskip 2mm}  
}
\makeatother

\definecolor{lightyellow}{HTML}{ffe599}
\definecolor{green}{HTML}{34a853}
\definecolor{lightcornflowerblue}{HTML}{c9daf8}
\definecolor{darkyellow}{rgb}{0.85, 0.65, 0.13}

\newcommand{\method}[0]{\texttt{Trust-Align}}
\newcommand{\NewMethod}[0]{\texttt{Ground-GRPO}}
\newcommand{\metric}[0]{\texttt{Trust-Score}}

\definecolor{nmcolor}{RGB}{255, 25, 26}

\definecolor{rowgray}{gray}{0.94}        
\definecolor{rowblue}{RGB}{235,241,250}  
\definecolor{rowgreen}{RGB}{240,250,240} 
\definecolor{rowrose}{RGB}{250,240,245}  
\definecolor{rowbeige}{RGB}{250,250,240} 
\definecolor{rowpeach}{RGB}{253,245,230} 

\newcolumntype{L}{>{\cellcolor{white}}l}

\begin{document}

\maketitle

\begin{abstract}
Generating grounded and trustworthy responses remains a key challenge for large language models (LLMs). While retrieval-augmented generation (RAG) with citation-based grounding holds promise, instruction-tuned models frequently fail even in straightforward scenarios: missing explicitly stated answers, citing incorrectly, or refusing when evidence is available. In this work, we explore how reinforcement learning (RL) and internal reasoning can enhance grounding in LLMs. We use the GRPO (Group Relative Policy Optimization) method to train models using verifiable outcome-based rewards targeting answer correctness, citation sufficiency, and refusal quality, without requiring gold reasoning traces or expensive annotations. Through comprehensive experiments across ASQA, QAMPARI, ELI5, and ExpertQA we show that reasoning-augmented models significantly outperform instruction-only variants, especially in handling unanswerable queries and generating well-cited responses. A two-stage training setup, first optimizing answer and citation behavior and then refusal, further improves grounding by stabilizing the learning signal. Additionally, we revisit instruction tuning via GPT-4 distillation and find that combining it with GRPO enhances performance on long-form, generative QA tasks. Overall, our findings highlight the value of reasoning, stage-wise optimization, and outcome-driven RL for building more verifiable and reliable LLMs.

\end{abstract}

\section{Introduction}

The deployment of large language models (LLMs) in information-seeking applications has grown rapidly, yet ensuring that these models generate factually reliable and grounded outputs remains an open challenge. A core concern is the problem of \textit{hallucination}, where models produce fluent but fabricated or unfaithful content \cite{Ji_2023}. Retrieval-Augmented Generation (RAG) offers a promising solution by conditioning responses on external documents, which improves factual accuracy and enables verification \cite{asai2023selfraglearningretrievegenerate}. Despite this, even advanced models like GPT-4 often mishandle retrieved information—leading to RAG-specific hallucinations when evidence is misused, omitted, or over-cited \cite{gao2023enablinglargelanguagemodels, song2025measuringenhancingtrustworthinessllms}.

Recently, several works have begun integrating retrieval more tightly into internal reasoning processes. For example, Search-o1 and RAG-Star embed retrieval into multi-step reasoning pipelines to dynamically acquire and filter supporting evidence \cite{li2025searcho1agenticsearchenhancedlarge, jiang2024ragstarenhancingdeliberativereasoning}. Other approaches like ReSearch and R1-Searcher train LLMs to autonomously issue retrieval queries using reinforcement learning (RL) \cite{chen2025researchlearningreasonsearch, song2025r1searcherincentivizingsearchcapability}, while CoT-RAG enhances chain-of-thought prompting with sub-query decomposition and knowledge graph guidance \cite{li2025cotragintegratingchainthought}. While these methods improve retrieval-driven reasoning, their primary focus is on complex QA or planning tasks and not directly on grounding alignment, such as accurate citation or proper refusal when no answer is found.

In this paper, we focus specifically on improving \emph{grounded response quality} in LLMs—ensuring that answers are correct, verifiably cited, and appropriately refused when evidence is lacking. We propose \NewMethod{}, a two-stage reinforcement learning framework based on Group Relative Policy Optimization (GRPO). We propose a novel hierarchial reward system that uses verifiable outcome-based rewards to enhance grounding in 3 directions: Answer Correctness, Citation Quality and Refusal.

\noindent \NewMethod{} proceeds in two stages:
\begin{enumerate}
\item \textbf{Answer and Citation Optimization:} The model is trained on answerable examples using rewards that encourage correct answers and minimal yet sufficient citations.
\item \textbf{Refusal Learning:} The model is then trained on a mixture of answerable and unanswerable examples, using a refusal-specific reward to discourage unsupported answers.
\end{enumerate}

Our results show that large reasoning models outperform instruction-tuned variants across multiple QA benchmarks—ASQA, QAMPARI, and ELI5—under the TrustScore metric \cite{song2025measuringenhancingtrustworthinessllms}, with notable gains in identifying unanswerable questions and providing precise, grounded citations. We find that reasoning-enhanced models benefit significantly more from \NewMethod{} than their instruction-only counterparts, highlighting the synergy between intermediate reasoning and reinforcement learning. Additionally, prior work has shown that GPT-4-based distillation improves model groundedness and citation quality. We find that distillation performs well in structured, list-style tasks like QAMPARI, while \NewMethod{} achieves stronger results on open-ended, long-form QA tasks such as ELI5, where generating coherent, well-cited explanations is more challenging.

\begin{AIbox}{Key Takeaways}
\begin{itemize}[leftmargin=1em]
\item \textbf{Reasoning boosts grounding:} Large Reasoning Models generate more grounded answers compared to instruct models.
\item \textbf{RL is more effective on reasoning models:} \NewMethod{} yields greater improvements on reasoning models than on instruction-tuned models.
\item \textbf{Staged training improves alignment:} Decomposing rewards across stages—first for answer quality, then for refusal—stabilizes training and leads to better overall grounding.
\item \textbf{Distillation and RL are complementary on longform QA datasets:} GPT-4 distillation and fine-tuning small models through SFT or DPO improves both the shortform and longform QA performance; GRPO further enhances grounding, especially in open-ended, long-form tasks.
\end{itemize}
\end{AIbox}

\begin{figure*}[h!]
    \centering
    \includegraphics[width=0.95\linewidth]{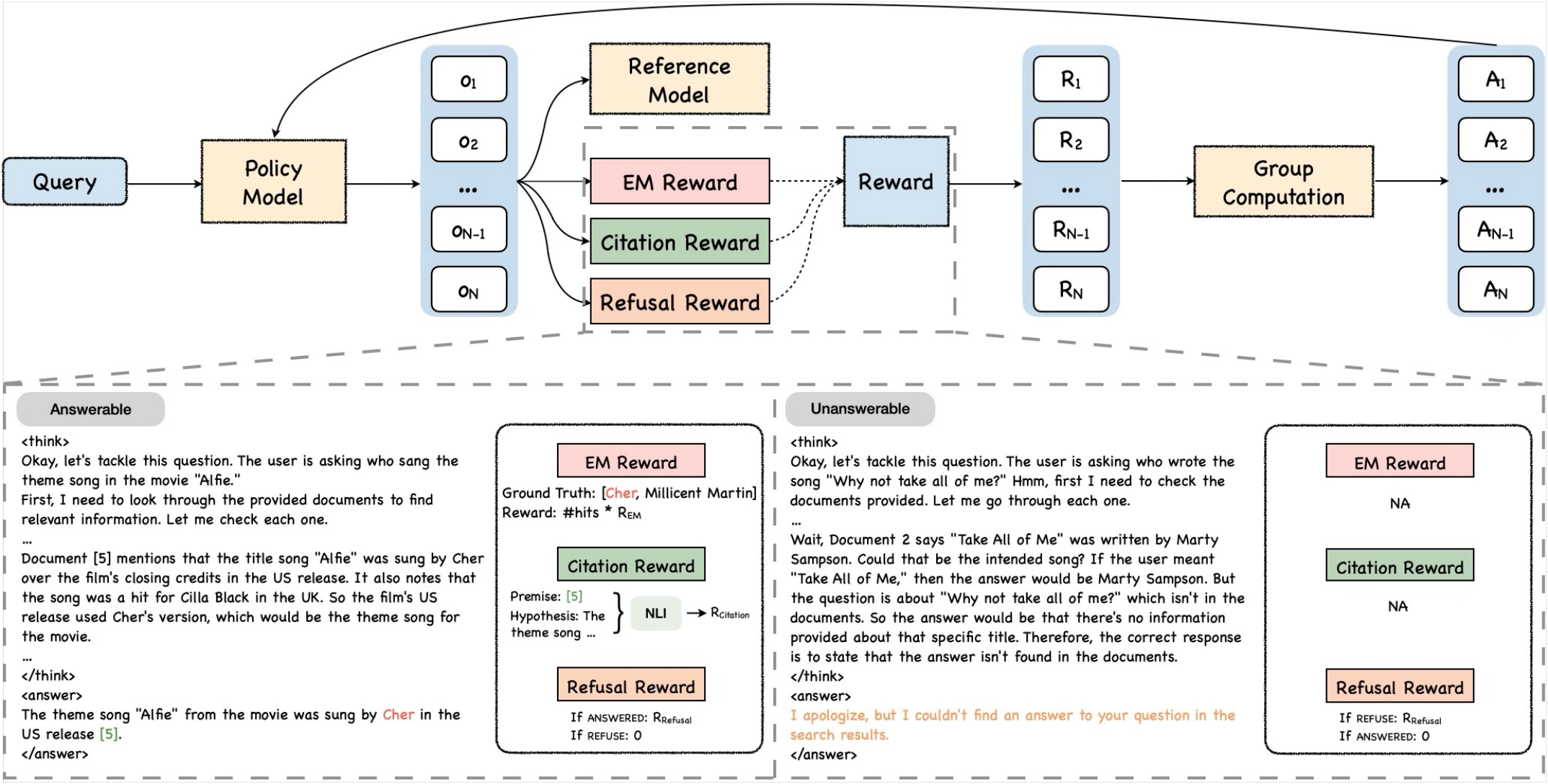}
    \caption{
Overview of the \NewMethod{} framework for reward-guided reinforcement learning in retrieval-augmented generation (RAG) settings. Given a user query, the Policy Model generates a set of candidate outputs $O_1, \ldots, O_N$, each of which is evaluated by a Reference Model to compute three distinct reward signals: (1) \textbf{EM Reward}, which checks for exact matches against ground-truth answers and provides a score based on the number of correct hits; (2) \textbf{Citation Reward}, which uses natural language inference (NLI) to verify whether the cited documents support the predicted answer; and (3) \textbf{Refusal Reward}, which incentivizes the model to appropriately abstain from answering unanswerable questions based on the evidence provided. The computed rewards $R_1, \ldots, R_N$ are aggregated using Group Computation to update the policy via reinforcement learning. The lower section of the figure illustrates examples for answerable and unanswerable queries, showing the model’s internal reasoning, selected answer, and corresponding reward computations. For answerable queries, both EM and citation rewards are used, while for unanswerable queries, a correct refusal is rewarded. \NewMethod{} thus encourages grounded, precise, and abstaining behavior in RAG systems.
}

    \label{fig:framework}
\end{figure*}

\section{Problem Definition and Method}
Given a question \(Q\) and a set of relevant documents \(D\), the LLM is instructed to generate a response \(S\) consisting of a set of citation-grounded statements \(\{s_1,s_2,...\}\). Each statement \(s_i\) is accompanied by a set of inline citations \(C_i = \{c_{i,1},c_{i,2},..\}\) that refers to the documents in \(D\), e.g., \texttt{statement1 [1][2] statement2 [3]}. If the relevant documents $D$ is insufficient to answer the question, the question is deemed \textit{unanswerable}, and the gold response $S$ would be a refusal statement, such as, \textit{``I apologize, but I couldn’t find an answer to your question in the search results''}. Otherwise, the question is \emph{answerable}.

\subsection{Two-stage Reinforcement Learning}
To facilitate robust learning of training signals by the LLM, we propose a two-stage RL approach that incorporates verifiable rewards. In Stage 1, the model is trained to accurately answer \textit{answerable} questions with appropriate citations while adhering to specified output format constraints. In Stage 2, the model is further trained on a mixture of answerable and unanswerable questions, enabling it to provide cited responses when possible and appropriately refuse to answer when questions are unanswerable.

\subsubsection{Stage 1.}
The model is trained only on \textit{answerable} questions in this stage. The reward function consists of three components: exact match (EM), citation, and format rewards. The primary objective is for the model to learn how to effectively utilize the provided documents to generate accurate responses to answerable questions.

The EM (Exact Match) reward is computed at the level of individual statements $s_i$. It is defined as follows:
\begin{align}
R_{\text{EM}} = 
\begin{cases}
0.5 & \scriptsize\text{if statement $s_i$ contains EM}, \\
0 & \scriptsize\text{otherwise} 
\end{cases}
\end{align}
This formulation assigns a reward of 0.5 for each statement that is deemed correct, and no reward for incorrect statements.

The citation reward is computed at the level of each correct EM statement $s_i$. It it defined as follows:
\begin{align}
R_{\text{citation}} = 
\begin{cases}
0.5 & \scriptsize\text{if statement } s_i \text{ containing EM has correct citation}, \\
-0.5 & \scriptsize\text{if statement } s_i \text{ containing EM has incorrect citation}, \\
0 & \scriptsize\text{otherwise}
\end{cases}
\end{align}
This encourages the model not only to produce accurate statements but also to support them with appropriate citations, while penalizing unsupported or incorrectly cited statements.

We further enforce the output format of the model's response where the thinking process should be enclosed within <think>...</think> tags, while the final answer only must appear within <answer>...</answer> tags. The <answer> section should not include any reasoning or justification. This constraint is enforced through a hard formatting reward, $R_{\text{format}}$, and a soft tag count reward, $R_{\text{tag\_count}}$, defined as follows:

\vspace{2mm}
{\small Let $\mathcal{T} = \{ \text{"<think>"}, \text{"</think>"}, \text{"<answer>"}, \text{"</answer>"} \}$.}

\begin{align}
    R_{\text{tag\_count}} = \frac{1}{|\mathcal{T}|} \sum_{t \in \mathcal{T}} \mathbf{1}_{\{\text{count}(t) = 1\}}
\end{align}

\begin{align}
    R_{\text{format}} = 
    \begin{cases}
        1.0 & \scriptsize\text{if output format is correct,}  \\
        0 & \scriptsize\text{otherwise} \\
    \end{cases}
\end{align}

Importantly, if output format is incorrect, the sample will not receive rewards for either \(R_{\text{EM}}\) or \(R_{\text{citation}}\). In such cases, the final reward defaults to \(R_{\text{tag\_count}}\) alone. The overall reward of Stage-1 is computed as the sum of the EM, citation, and formatting rewards:
\begin{align}
    R_\text{Stage-1} & = R_\text{EM} + R_\text{citation} + R_\text{tag\_count} + R_\text{format} 
\end{align}

\subsubsection{Stage-2.}
In Stage-2, the model is trained on a mix of answerable and unanswerable questions. This stage introduces a refusal reward, $R_{\text{refusal}}$, to teach the model to identify when provided documents lack sufficient information to answer a question.

\begin{align}
    R_{\text{refusal}} = 
    \begin{cases}
        0 & {\scriptsize \text{if \texttt{answerable} and } r_{score} > 0.85} \\
        0.5 & {\scriptsize \text{if \texttt{answerable} and } r_{score} < 0.85} \\
        r_{score} & {\scriptsize \text{if \texttt{unanswerable} and } r_{score} > 0.85} \\
        0 & {\scriptsize \text{if \texttt{unanswerable} and } r_{score} < 0.85} \\
    \end{cases}
\end{align}

where $r_{score}$ refers to the refusal score that is calculated using a fuzzy matching algorithm, producing a value between 0 and 1, with values closer to 1 indicating a higher similarity to the gold standard refusal response. The final reward for Stage-2 is the sum of Stage-1 Rewards with the addition of $R_{\text{refusal}}$. 

The hierarchical rewards can be visualized in \Cref{algo:main}. 

\begin{algorithm}[h]
\fontsize{9.6pt}{10.8pt}\selectfont
\caption{Hierarchical Rewards}\label{algo:main}
\begin{algorithmic}[1]
\Require Policy Model $\mathcal{M}$
\State \textbf{Input:} Questions \(Q\), Task instruction \(I\), Documents \(D\), Generated response \(S\)
\State Initialize reward \(R = 0\)
\State \texttt{\(R\) += \(R_{\text{tag\_count}}\)}
\If{\(R_{\text{format}} = 1\) and \(R_{\text{tag\_count}} = 1\)}
    \State \texttt{\(R\) += \(R_{\text{format}}\)}
    \If{is\_answerable(\(Q, D\))}
        \If{\(r_{score}(S) < 0.85\)}
            \State \texttt{\(R\) += \(R_{refusal}\)}
            \State \texttt{\(R\) += \(R_{\text{correctness}}\)}
        \EndIf
    \ElsIf{not is\_answerable(\(Q, D\))}
        \State \texttt{\(R\) += \(R_{refusal}\)}        
    \EndIf
\EndIf
\State \textbf{Return} \(R\)
\end{algorithmic}
\end{algorithm}

\begin{figure*}[ht!]
    \centering
    \includegraphics[width=\textwidth]{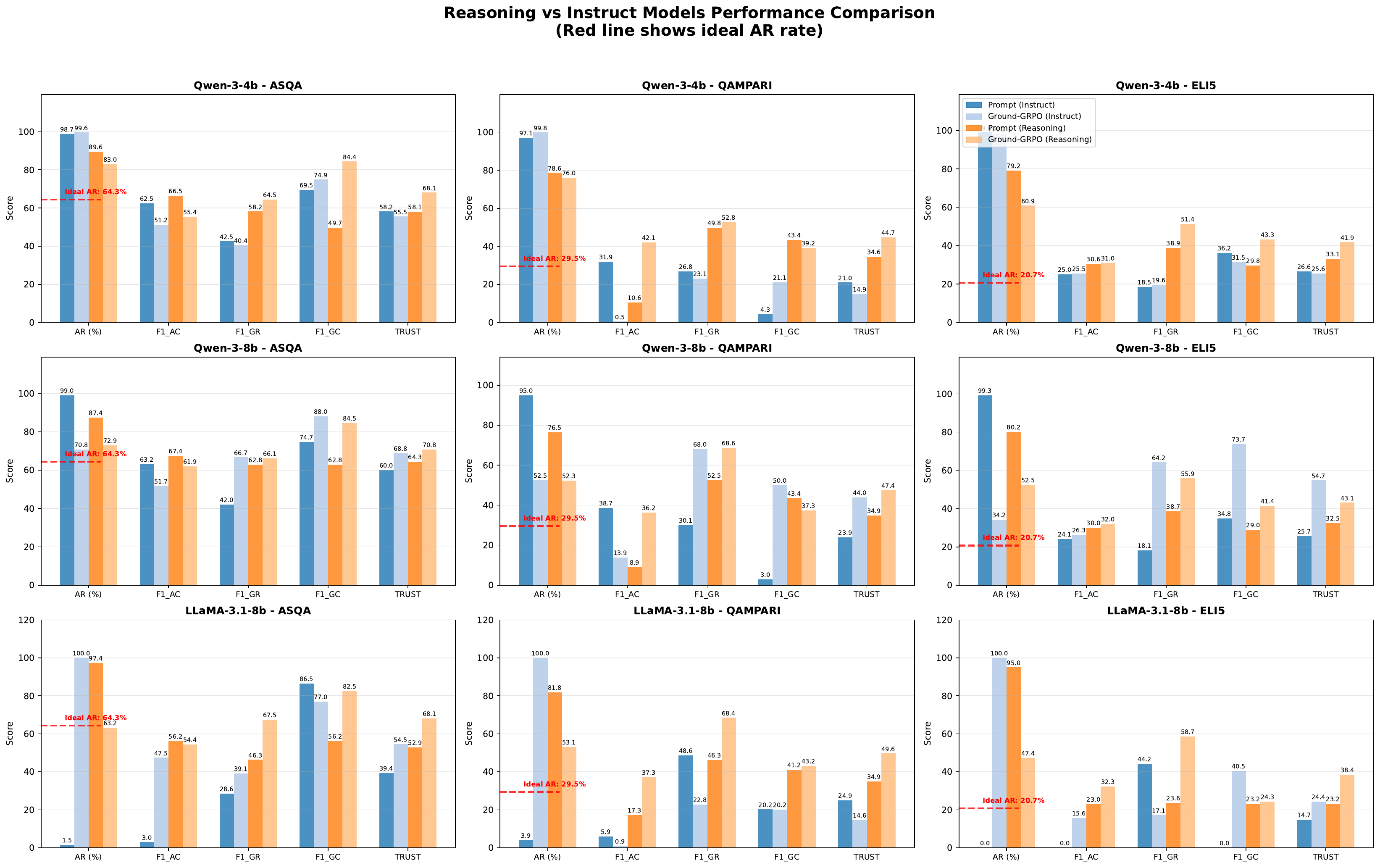}
    \caption{Performance comparison between reasoning and instruct models across three model families (Qwen-3-4b, Qwen-3-8b, LLaMA-3.1-8b) and three datasets (ASQA, QAMPARI, ELI5). Four configurations are compared: Prompt (Instruct) in dark blue, \NewMethod{} (Instruct) in light blue, Prompt (Reasoning) in dark orange, and \NewMethod{} (Reasoning) in light orange. Reasoning models consistently outperform their instruct counterparts on grounding metrics (\textbf{F1$_{\text{GR}}$}) and trust scores (\textbf{TRUST}), demonstrating superior reliability in handling answerable and unanswerable questions. The \NewMethod{} training objective further enhances performance for both instruct and reasoning variants, with \NewMethod{} (Reasoning) achieving the best overall results across most metrics. Notable improvements are observed in grounded refusals and trust scores, validating the effectiveness of reasoning capabilities for responsible question answering. \textbf{F1$_{\text{AC}}$} := Answer Correctness F1; \textbf{F1$_{\text{GR}}$} := Grounded Refusals F1; \textbf{F1$_{\text{GC}}$} := Grounded Citations F1; \textbf{TRUST} := \metric{}.}
    \label{fig:reasoning-vs-instruct}
\end{figure*}

\section{Experimental Setup}

\subsubsection{Training Data}
We train our models using samples from the Trust-Align SFT dataset, which is constructed using questions from the ASQA, QAMPARI, and ELI5 datasets, paired with documents retrieved from web sources for each question.

For Stage 1 training, we select 100 answerable questions from each of the three datasets, focusing solely on samples where the answer can be found within the provided documents. For Stage 2 training, we use a larger dataset of 1,000 samples per subset (3,000 total), composed of an equal mix of answerable and unanswerable questions.

\subsection{Training Algorithm}
We train our models using the GRPO algorithm, as implemented in the TRL library. Each training sample is associated with 8 generated responses, and training is conducted for 4 epochs. We use a global batch size of 384, a learning rate of $1.0 \times 10^{-5}$, a cosine learning rate scheduler, and a warm-up ratio of 0.1.

\subsection{Models and Baselines}

Our main experiments use three backbone models: LLaMA3.1-8B, Qwen3-4B, and Qwen3-8B \cite{grattafiori2024llama3, qwen3technicalreport}. These models were selected because both reasoning and instruction-tuned variants are publicly available. Notably, the Qwen3 series supports hybrid generation modes—with and without explicit reasoning—making it suitable for studying the effects of reasoning supervision. For LLaMA3.1-8B, we use the DeepSeek R1 distilled version as the reasoning-tuned variant \cite{deepseekai2025deepseekr1incentivizingreasoningcapability}. To further strengthen the generality of our findings, we include additional results using the Gemma2-9B model for instruction-tuned experiments \cite{gemma_2024}.

As baselines, we adopt the \method{} SFT and DPO datasets, which consist of approximately 10K and 17K samples, respectively. These externally sourced, distilled preference datasets train instruction-tuned models and serve as benchmarks against which we compare our proposed \NewMethod{}. In contrast to previous work that hasn’t leveraged reasoning-enhanced models for groundedness, our baseline comparisons focus exclusively on instruction-tuned variants. The \method{} dataset offers a distilled, high-quality alignment resource comprising about 19K carefully curated examples. Each sample pairs a “preferred” response—grounded in retrieved documents with proper citations—with an “unpreferred” counterpart, which may exhibit hallucination, mis-citation, over-responsiveness, or unwarranted refusal. Data was constructed by sampling questions from benchmarks like ASQA, QAMPARI, and ELI5; retrieving supporting documents; and generating responses (both positive and negative) using strong models such as GPT‑4. The resulting \method{} dataset is used by \citet{song2025measuringenhancingtrustworthinessllms} to align LLMs for grounded response generation using SFT and DPO methods. We consider \method{} baselines as distillation approaches (~\Cref{fig:trust-align}).

\subsection{Evaluation Benchmark}
For evaluation, we use the Trust Score test set. The benchmark was constructed using the test splits of the ASQA, ELI5, and QAMPARI datasets, following the same method as the Trust-Align dataset. To assess the models’ robustness, the benchmark also includes samples from ExpertQA, enabling evaluation on out-of-domain (OOD) questions

\subsection{Evaluation Metrics}
Following \citeauthor{song2025measuringenhancingtrustworthinessllms}, we use the \metric{} metric to evaluate the overall groundedness of model responses. \metric{} is computed as the average of three key components:
\begin{enumerate}
    \item \textbf{Answer Correctness} (F1\textsubscript{AC}): Assesses the factual correctness of the model's answer by comparing it against a list of ground-truth answers (can have more than one per question).

    \item \textbf{Grounded Refusals} (F1\textsubscript{GR}): Measures the model’s ability to detect when a question is unanswerable based on the provided documents and to appropriately refuse to answer.

    \item \textbf{Grounded Citations} (F1\textsubscript{GC}): Evaluates the quality of citations by measuring the precision and recall of evidence linked to statements in the response. Responses are penalized for including unsupported claims, redundant citations, or failing to cite necessary evidence.

\end{enumerate}
\begin{figure*}[t!]
    \centering
    \includegraphics[width=\textwidth]{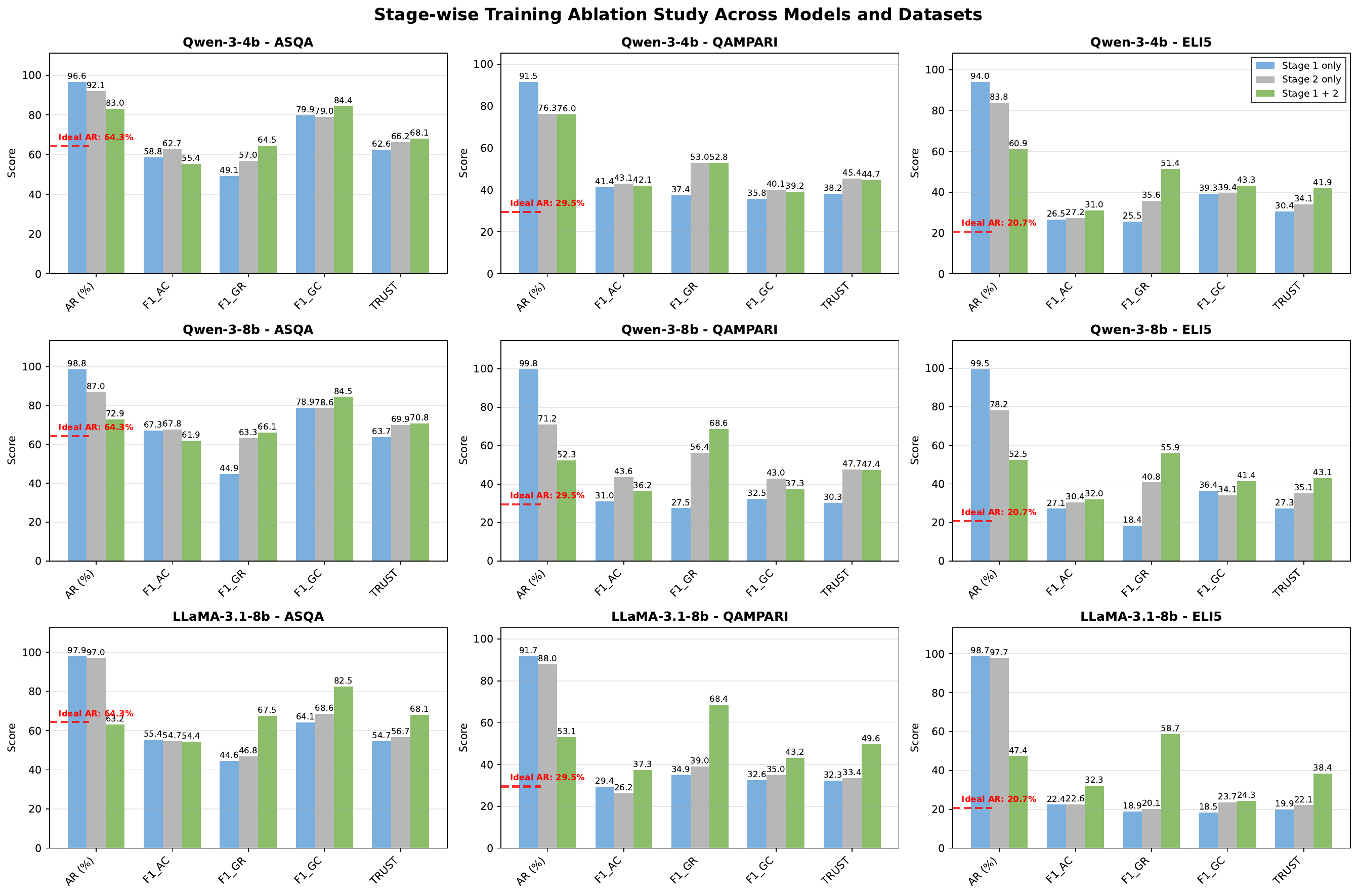}
    \caption{Stage-wise training ablation study across three models (Qwen-3-4b, Qwen-3-8b, LLaMA-3.1-8b) and three datasets (ASQA, QAMPARI, ELI5). Each subplot shows performance comparison between three training strategies: Stage 1 only (blue), Stage 2 only (gray), and Stage 1 + 2 combined (green). Stage 1 + 2 training consistently improves grounding metrics (\textbf{F1$_{\text{GR}}$}) across all model-dataset combinations. The combined training approach achieves the highest trust scores (\textbf{TRUST}) in most cases, indicating better overall reliability.}
    \label{fig:stagewise-ablation}
\end{figure*}
Additionally, we report the \emph{Answer Ratio} (AR), which measures the proportion of questions answered by the model. Although we do not directly compare models based on this metric, we expect models to align with the ideal AR for each dataset. The AR is computed as:
\[
\text{AR} = \frac{\# \text{questions answered}}{\# \text{total questions in the dataset}}
\]
Large deviations from the ideal AR may indicate that a model is either overly responsive or overly cautious in answering questions. Note that this behavior of the model is captured in F1$_{GR}$.

\noindent The detailed formulas for metrics can be found in the \Cref{sec:metrics_app}.

\section{Results}

\subsection{Reasoning Promotes Grounded Responses}
\Cref{fig:reasoning-vs-instruct} demonstrates that reasoning models\footnote{Models with internal thought processes trained to produce high-quality reasoning traces.} consistently outperform their instruct counterparts across the ASQA, QAMPARI, and ELI5 datasets. Specifically, when prompted in a zero-shot setting (i.e., without any additional fine-tuning), reasoning models achieve higher \metric{} scores in 8 out of 9 configurations. This performance gap is still present even after training with \NewMethod{}.

A deeper examination of the individual components of \metric{} reveals that the most substantial gains come from improvements in \textbf{F1$_{\text{GR}}$}. This suggests that reasoning enhances the model’s ability to identify unanswerable questions based on provided documents and to refuse appropriately. Notably, this pattern holds across both prompting ($\uparrow 13.1\%$) and GRPO ($\uparrow 20.4\%$) across all configurations, emphasizing the general utility of explicit reasoning in promoting more grounded responses.

\begin{AIbox}{TL;DR of Reasoning Promotes Grounded Responses}
\begin{itemize}[leftmargin=1em]
\item Reasoning improves the Groundedness of Models across different datasets  
\item Reasoning is especially helpful in enhancing the model's ability to identify and refuse unanswerable questions
\end{itemize}
\end{AIbox}
\begin{table*}[htb!]
\caption{Results of applying \NewMethod{} to distilled versions of LLaMA-3.1, Qwen-3, and Gemma-2 models on the ASQA, QAMPARI, and ELI5 datasets. The distillation is performed using SFT or DPO on the \texttt{Trust-Align} dataset, itself a distilled dataset from GPT-4. \NewMethod{} is applied to these distilled variants, and the results show that it is beneficial only for longform QA datasets such as ELI5.
}
\label{table:Trust-Align}

\centering
\resizebox{\textwidth}{!}{
\begin{tabular}{ll*{16}{c}}
\toprule
\multirow{2}{*}{\textbf{Model}} & \multirow{2}{*}{\textbf{Type}} & \multicolumn{5}{c}{\textbf{ASQA} \textit{(610 answerable, 338 unanswerable)}} & \multicolumn{5}{c}{\textbf{QAMPARI} \textit{(295 answerable, 705 unanswerable)}} & \multicolumn{5}{c}{\textbf{ELI5} \textit{(207 answerable, 793 unanswerable)}} \\

\cmidrule(lr){3-7}\cmidrule(lr){8-12}\cmidrule(lr){13-17}

& & \textbf{AR (\%)} & \textbf{F1$_{\text{AC}}$} & \textbf{F1$_{\text{GR}}$} & \textbf{F1$_{\text{GC}}$} & \textbf{TRUST} & \textbf{AR (\%)} & \textbf{F1$_{\text{AC}}$} & \textbf{F1$_{\text{GR}}$} & \textbf{F1$_{\text{GC}}$} & \textbf{TRUST} & \textbf{AR (\%)} & \textbf{F1$_{\text{AC}}$} & \textbf{F1$_{\text{GR}}$} & \textbf{F1$_{\text{GC}}$} & \textbf{TRUST}\\
\midrule

\multirow{4}{*}{\mystack{Qwen-3}{-4b}} 
 & \NewMethod{} (Instruct) & 99.58 & 51.25 & 40.42 & 74.91 & 55.53 & 99.80 & 0.46 & 23.10 & 21.15 & 14.90 & 98.20 & 25.51 & 19.63 & 31.52 & 25.55 \\
& \cellcolor{rowgray}SFT & \cellcolor{rowgray}72.78 & \cellcolor{rowgray}46.88 & \cellcolor{rowgray}61.34 & \cellcolor{rowgray}75.76 & \cellcolor{rowgray}61.33 & \cellcolor{rowgray}32.73 & \cellcolor{rowgray}45.41 & \cellcolor{rowgray}71.61 & \cellcolor{rowgray}40.78 & \cellcolor{rowgray}52.60 & \cellcolor{rowgray}34.90 & \cellcolor{rowgray}16.31 & \cellcolor{rowgray}58.65 & \cellcolor{rowgray}34.66 & \cellcolor{rowgray}36.54 \\
& \cellcolor{rowgray}\(\text{SFT} \to \NewMethod{}\) & \cellcolor{rowgray}68.04 & \cellcolor{rowgray}47.22 & \cellcolor{rowgray}62.88 & \cellcolor{rowgray}81.94 & \cellcolor{rowgray}64.01 & \cellcolor{rowgray}57.00 & \cellcolor{rowgray}29.83 & \cellcolor{rowgray}61.60 & \cellcolor{rowgray}36.93 & \cellcolor{rowgray}42.79 & \cellcolor{rowgray}29.00 & \cellcolor{rowgray}20.32 & \cellcolor{rowgray}61.85 & \cellcolor{rowgray}51.71 & \cellcolor{rowgray}44.62 \\
  & \cellcolor{rowgray}\(\NewMethod{} \to \text{SFT}\) &\cellcolor{rowgray} 75.11 &\cellcolor{rowgray} 44.68 &\cellcolor{rowgray} 55.28 &\cellcolor{rowgray} 71.69 &\cellcolor{rowgray} 57.22 &\cellcolor{rowgray} 12.30 &\cellcolor{rowgray} 22.97 &\cellcolor{rowgray} 57.66 &\cellcolor{rowgray} 38.00 &\cellcolor{rowgray} 39.54 &\cellcolor{rowgray} 41.70 &\cellcolor{rowgray} 11.65 &\cellcolor{rowgray} 50.85 &\cellcolor{rowgray} 22.73 &\cellcolor{rowgray} 28.41 \\
 \cdashlinelr{2-17}
 & \cellcolor{rowpeach}\(\text{SFT} \to \text{DPO}\) & \cellcolor{rowpeach}65.61 & \cellcolor{rowpeach}50.25 & \cellcolor{rowpeach}64.77 & \cellcolor{rowpeach}79.60 & \cellcolor{rowpeach}64.88 & \cellcolor{rowpeach}29.20 & \cellcolor{rowpeach}37.82 & \cellcolor{rowpeach}71.67 & \cellcolor{rowpeach}42.99 & \cellcolor{rowpeach}50.83 & \cellcolor{rowpeach}26.80 & \cellcolor{rowpeach}20.49 & \cellcolor{rowpeach}61.76 & \cellcolor{rowpeach}38.01 & \cellcolor{rowpeach}40.09 \\
  & \cellcolor{rowpeach}\(\text{SFT} \to \text{DPO} \to \NewMethod{}\) & \cellcolor{rowpeach}48.31 & \cellcolor{rowpeach}44.00 & \cellcolor{rowpeach}61.84 & \cellcolor{rowpeach}86.79 & \cellcolor{rowpeach}64.21 & \cellcolor{rowpeach}46.80 & \cellcolor{rowpeach}39.32 & \cellcolor{rowpeach}68.32 & \cellcolor{rowpeach}44.00 & \cellcolor{rowpeach}50.55 & \cellcolor{rowpeach}20.70 & \cellcolor{rowpeach}20.29 & \cellcolor{rowpeach}64.36 & \cellcolor{rowpeach}59.18 & \cellcolor{rowpeach}47.94 \\
  & \cellcolor{rowpeach}\(\NewMethod{} \to \text{SFT} \to \text{DPO}\) &\cellcolor{rowpeach} 67.62 &\cellcolor{rowpeach} 45.55 &\cellcolor{rowpeach} 54.06 &\cellcolor{rowpeach} 75.74 &\cellcolor{rowpeach} 58.45 &\cellcolor{rowpeach} 15.20 &\cellcolor{rowpeach} 18.79 &\cellcolor{rowpeach} 57.50 &\cellcolor{rowpeach} 39.02 &\cellcolor{rowpeach} 38.44 &\cellcolor{rowpeach} 37.40 &\cellcolor{rowpeach} 12.97 &\cellcolor{rowpeach} 51.85 &\cellcolor{rowpeach} 24.68 &\cellcolor{rowpeach} 29.83 \\
\midrule

\multirow{4}{*}{\mystack{Qwen-3}{-8b}} 
 & \NewMethod{} (Instruct) & 70.78 & 51.73 & 66.67 & 87.97 & 68.79 & 52.50 & 13.90 & 67.96 & 49.98 & 43.95 & 34.20 & 26.29 & 64.22 & 73.71 & 54.74 \\
  & \cellcolor{rowgray}SFT & \cellcolor{rowgray}67.72 & \cellcolor{rowgray}63.86 & \cellcolor{rowgray}66.37 & \cellcolor{rowgray}80.71 & \cellcolor{rowgray}70.31 & \cellcolor{rowgray}29.50 & \cellcolor{rowgray}56.61 & \cellcolor{rowgray}75.00 & \cellcolor{rowgray}50.85 & \cellcolor{rowgray}60.82 & \cellcolor{rowgray}15.80 & \cellcolor{rowgray}25.84 & \cellcolor{rowgray}64.31 & \cellcolor{rowgray}38.68 & \cellcolor{rowgray}42.95 \\
  & \cellcolor{rowgray}\(\text{SFT} \to \NewMethod{}\) & \cellcolor{rowgray}66.14 & \cellcolor{rowgray}52.37 & \cellcolor{rowgray}66.86 & \cellcolor{rowgray}89.32 & \cellcolor{rowgray}69.52 & \cellcolor{rowgray}50.50 & \cellcolor{rowgray}44.00 & \cellcolor{rowgray}67.50 & \cellcolor{rowgray}40.78 & \cellcolor{rowgray}50.76 & \cellcolor{rowgray}38.80 & \cellcolor{rowgray}25.21 & \cellcolor{rowgray}63.99 & \cellcolor{rowgray}53.38 & \cellcolor{rowgray}47.53 \\
& \cellcolor{rowgray}\(\NewMethod{} \to \text{SFT}\) &\cellcolor{rowgray} 84.81 &\cellcolor{rowgray} 49.70 &\cellcolor{rowgray} 57.16 &\cellcolor{rowgray} 72.92 &\cellcolor{rowgray} 59.93 &\cellcolor{rowgray} 22.60 &\cellcolor{rowgray} 33.40 &\cellcolor{rowgray} 66.91 &\cellcolor{rowgray} 39.27 &\cellcolor{rowgray} 46.52 &\cellcolor{rowgray} 37.80 &\cellcolor{rowgray} 14.25 &\cellcolor{rowgray} 53.01 &\cellcolor{rowgray} 25.36 &\cellcolor{rowgray} 30.87 \\

   \cdashlinelr{2-17}
 & \cellcolor{rowpeach}\(\text{SFT} \to \text{DPO}\) & \cellcolor{rowpeach}52.43 & \cellcolor{rowpeach}63.22 & \cellcolor{rowpeach}64.94 & \cellcolor{rowpeach}76.97 & \cellcolor{rowpeach}68.38 & \cellcolor{rowpeach}31.90 & \cellcolor{rowpeach}51.79 & \cellcolor{rowpeach}75.09 & \cellcolor{rowpeach}53.20 & \cellcolor{rowpeach}60.03 & \cellcolor{rowpeach}7.90 & \cellcolor{rowpeach}21.68 & \cellcolor{rowpeach}58.79 & \cellcolor{rowpeach}41.97 & \cellcolor{rowpeach}40.81 \\
  & \cellcolor{rowpeach}\(\text{SFT} \to \text{DPO} \to \NewMethod{}\) &  \cellcolor{rowpeach}44.09 & \cellcolor{rowpeach}47.60 & \cellcolor{rowpeach}63.03 & \cellcolor{rowpeach}94.92 & \cellcolor{rowpeach}68.52 & \cellcolor{rowpeach}47.20 & \cellcolor{rowpeach}42.76 & \cellcolor{rowpeach}70.29 & \cellcolor{rowpeach}47.05 & \cellcolor{rowpeach}53.37 & \cellcolor{rowpeach}29.40 & \cellcolor{rowpeach}25.35 & \cellcolor{rowpeach}67.64 & \cellcolor{rowpeach}62.86 & \cellcolor{rowpeach}51.95 \\
& \cellcolor{rowpeach}\(\NewMethod{} \to \text{SFT} \to \text{DPO}\) &\cellcolor{rowpeach} 77.32 &\cellcolor{rowpeach} 52.11 &\cellcolor{rowpeach} 58.00 &\cellcolor{rowpeach} 80.49 &\cellcolor{rowpeach} 63.54 &\cellcolor{rowpeach} 19.70 &\cellcolor{rowpeach} 20.33 &\cellcolor{rowpeach} 65.77 &\cellcolor{rowpeach} 43.49 &\cellcolor{rowpeach} 43.19 &\cellcolor{rowpeach} 32.80 &\cellcolor{rowpeach} 14.02 &\cellcolor{rowpeach} 55.22 &\cellcolor{rowpeach} 27.64 &\cellcolor{rowpeach} 32.29 \\
  
\midrule
  
\multirow{4}{*}{\mystack{LLaMA-3.1}{-8b}} 
   & \NewMethod{} (Instruct) & 100.00 & 47.46 & 39.15 & 77.02 & 54.54 & 100.00 & 0.93 & 22.78 & 20.20 & 14.63 & 100.00 & 15.63 & 17.15 & 40.47 & 24.42 \\
    & \cellcolor{rowgray}SFT & \cellcolor{rowgray}72.15 & \cellcolor{rowgray}53.52 & \cellcolor{rowgray}64.71 & \cellcolor{rowgray}82.80 & \cellcolor{rowgray}67.01 & \cellcolor{rowgray}33.70 & \cellcolor{rowgray}54.75 & \cellcolor{rowgray}72.24 & \cellcolor{rowgray}47.26 & \cellcolor{rowgray}58.08 & \cellcolor{rowgray}21.10 & \cellcolor{rowgray}21.77 & \cellcolor{rowgray}65.82 & \cellcolor{rowgray}44.81 & \cellcolor{rowgray}44.13 \\
  & \cellcolor{rowgray}\(\text{SFT} \to \NewMethod{}\) & \cellcolor{rowgray}56.65 & \cellcolor{rowgray}44.96 & \cellcolor{rowgray}65.24 & \cellcolor{rowgray}82.80 & \cellcolor{rowgray}64.33 & \cellcolor{rowgray}40.90 & \cellcolor{rowgray}34.38 & \cellcolor{rowgray}67.56 & \cellcolor{rowgray}40.93 & \cellcolor{rowgray}47.62 & \cellcolor{rowgray}34.00 & \cellcolor{rowgray}22.36 & \cellcolor{rowgray}63.89 & \cellcolor{rowgray}46.31 & \cellcolor{rowgray}44.19 \\
  & \cellcolor{rowgray}\(\NewMethod{} \to \text{SFT}\) & \cellcolor{rowgray}68.35 &\cellcolor{rowgray} 36.74 &\cellcolor{rowgray} 55.12 &\cellcolor{rowgray} 54.97 &\cellcolor{rowgray} 48.94 &\cellcolor{rowgray} 3.68 &\cellcolor{rowgray} 6.15 &\cellcolor{rowgray} 47.06 &\cellcolor{rowgray} 32.61 &\cellcolor{rowgray} 28.61 &\cellcolor{rowgray} 26.00 &\cellcolor{rowgray} 6.64 &\cellcolor{rowgray} 53.21 &\cellcolor{rowgray} 17.94 &\cellcolor{rowgray} 25.93 \\
   \cdashlinelr{2-17}
 & \cellcolor{rowpeach}\(\text{SFT} \to \text{DPO}\) & \cellcolor{rowpeach}33.65 & \cellcolor{rowpeach}45.48 & \cellcolor{rowpeach}58.95 & \cellcolor{rowpeach}91.94 & \cellcolor{rowpeach}65.46 & \cellcolor{rowpeach}28.10 & \cellcolor{rowpeach}49.31 & \cellcolor{rowpeach}73.42 & \cellcolor{rowpeach}55.04 & \cellcolor{rowpeach}59.26 & \cellcolor{rowpeach}4.60 & \cellcolor{rowpeach}17.26 & \cellcolor{rowpeach}56.79 & \cellcolor{rowpeach}69.73 & \cellcolor{rowpeach}47.93 \\
  & \cellcolor{rowpeach}\(\text{SFT} \to \text{DPO} \to \NewMethod{}\) & \cellcolor{rowpeach}31.22 & \cellcolor{rowpeach}36.13 & \cellcolor{rowpeach}56.45 & \cellcolor{rowpeach}87.50 & \cellcolor{rowpeach}60.03 & \cellcolor{rowpeach}44.60 & \cellcolor{rowpeach}35.63 & \cellcolor{rowpeach}66.02 & \cellcolor{rowpeach}40.39 & \cellcolor{rowpeach}47.35 & \cellcolor{rowpeach}17.70 & \cellcolor{rowpeach}18.14 & \cellcolor{rowpeach}62.29 & \cellcolor{rowpeach}54.52 & \cellcolor{rowpeach}44.98 \\
  & \cellcolor{rowpeach}\(\NewMethod{} \to \text{SFT} \to \text{DPO}\) &\cellcolor{rowpeach} 56.33 &\cellcolor{rowpeach} 36.56 &\cellcolor{rowpeach} 51.51 &\cellcolor{rowpeach}60.96 &\cellcolor{rowpeach} 49.68 &\cellcolor{rowpeach} 20.30 &\cellcolor{rowpeach} 20.08 &\cellcolor{rowpeach} 58.82 &\cellcolor{rowpeach} 27.74 &\cellcolor{rowpeach} 35.55 &\cellcolor{rowpeach} 15.40 &\cellcolor{rowpeach} 7.20 &\cellcolor{rowpeach} 52.85 &\cellcolor{rowpeach} 19.53 &\cellcolor{rowpeach} 26.53 \\

  \midrule
  
\multirow{4}{*}{\mystack{Gemma-2}{-9b}} 
   & \NewMethod{} (Instruct) &  100.00 & 50.01 & 39.15 & 80.19 & 56.45 & 100.00 & 35.98 & 22.78 & 22.21 & 26.99 & 100.00 & 14.83 & 17.15 & 48.35 & 26.78\\
     & \cellcolor{rowgray}SFT & \cellcolor{rowgray}75.32 & \cellcolor{rowgray}56.12 & \cellcolor{rowgray}66.95 & \cellcolor{rowgray}85.12 & \cellcolor{rowgray}69.40 & \cellcolor{rowgray}36.90 & \cellcolor{rowgray}52.71 & \cellcolor{rowgray}76.10 & \cellcolor{rowgray}50.58 & \cellcolor{rowgray}59.80 & \cellcolor{rowgray}25.20 & \cellcolor{rowgray}23.09 & \cellcolor{rowgray}65.08 & \cellcolor{rowgray}48.87 & \cellcolor{rowgray}45.68 \\
  & \cellcolor{rowgray}\(\text{SFT} \to \NewMethod{}\) & \cellcolor{rowgray}47.15 & \cellcolor{rowgray}49.20 & \cellcolor{rowgray}63.55 & \cellcolor{rowgray}89.48 & \cellcolor{rowgray}67.41 & \cellcolor{rowgray}49.30 & \cellcolor{rowgray}42.13 & \cellcolor{rowgray}66.91 & \cellcolor{rowgray}44.57 & \cellcolor{rowgray}51.20 & \cellcolor{rowgray}26.10 & \cellcolor{rowgray}21.58 & \cellcolor{rowgray}65.69 & \cellcolor{rowgray}64.37 & \cellcolor{rowgray}50.55 \\
  & \cellcolor{rowgray}\(\NewMethod{} \to \text{SFT}\) &\cellcolor{rowgray} 68.35 &\cellcolor{rowgray} 37.85 &\cellcolor{rowgray} 51.34 &\cellcolor{rowgray} 51.86 &\cellcolor{rowgray} 47.01 &\cellcolor{rowgray} 7.50 &\cellcolor{rowgray} 12.97 &\cellcolor{rowgray} 50.26 &\cellcolor{rowgray} 24.78 &\cellcolor{rowgray} 29.34 &\cellcolor{rowgray} 32.40 &\cellcolor{rowgray} 7.34 &\cellcolor{rowgray} 50.13 &\cellcolor{rowgray} 12.38 &\cellcolor{rowgray} 23.29 \\
   \cdashlinelr{2-17}
 & \cellcolor{rowpeach}\(\text{SFT} \to \text{DPO}\) & \cellcolor{rowpeach}62.45 & \cellcolor{rowpeach}57.04 & \cellcolor{rowpeach}69.54 & \cellcolor{rowpeach}88.96 & \cellcolor{rowpeach}71.85 & \cellcolor{rowpeach}33.30 & \cellcolor{rowpeach}47.45 & \cellcolor{rowpeach}75.86 & \cellcolor{rowpeach}55.13 & \cellcolor{rowpeach}59.48 & \cellcolor{rowpeach}14.80 & \cellcolor{rowpeach}24.32 & \cellcolor{rowpeach}63.53 & \cellcolor{rowpeach}67.11 & \cellcolor{rowpeach}51.65 \\
  & \cellcolor{rowpeach}\(\text{SFT} \to \text{DPO} \to \NewMethod{}\) &\cellcolor{rowpeach}46.73 & \cellcolor{rowpeach}48.60 & \cellcolor{rowpeach}65.51 & \cellcolor{rowpeach}89.69 & \cellcolor{rowpeach}67.93 & \cellcolor{rowpeach}49.70 & \cellcolor{rowpeach}41.92 & \cellcolor{rowpeach}67.18 & \cellcolor{rowpeach}39.29 & \cellcolor{rowpeach}49.46 & \cellcolor{rowpeach}25.50 & \cellcolor{rowpeach}21.50 & \cellcolor{rowpeach}65.38 & \cellcolor{rowpeach}69.90 & \cellcolor{rowpeach}52.26 \\
  & \cellcolor{rowpeach}\(\NewMethod{} \to \text{SFT} \to \text{DPO}\) &\cellcolor{rowpeach} 58.33 &\cellcolor{rowpeach} 45.90 &\cellcolor{rowpeach} 55.63 &\cellcolor{rowpeach} 55.01 &\cellcolor{rowpeach} 52.18 &\cellcolor{rowpeach} 6.40 &\cellcolor{rowpeach} 6.69 &\cellcolor{rowpeach} 48.91 &\cellcolor{rowpeach} 27.97 &\cellcolor{rowpeach} 27.85 &\cellcolor{rowpeach} 29.60 &\cellcolor{rowpeach} 13.85 &\cellcolor{rowpeach} 53.12 &\cellcolor{rowpeach} 14.09 &\cellcolor{rowpeach} 27.02 \\
  
\bottomrule
\end{tabular}
}
\end{table*}
\subsection{Training with RL Improves Groundedness}
\Cref{fig:reasoning-vs-instruct} further indicates that training with GRPO leads to substantial gains in groundedness across all model types and datasets. For each of the three models evaluated on ASQA, QAMPARI, and ELI5, \metric{} improves consistently following GRPO training. The gains are especially pronounced in reasoning models, which improve by an average of 11.5 points, compared to 7.0 points for their instruct counterparts.

A breakdown of the \metric{} subcomponents reveals differing patterns of improvement. For instruct models, the most notable gains are in \textbf{F1$_{\text{GC}}$} ($\uparrow 16.4\%$), suggesting that GRPO helps these models better justify their responses with minimally sufficient citations. In contrast, reasoning models exhibit broad improvements across all three submetrics: \textbf{F1$_{\text{AC}}$} ($\uparrow 8.0\%$), \textbf{F1$_{\text{GR}}$} ($\uparrow 15.2\%$), and \textbf{F1$_{\text{GC}}$} ($\uparrow 11.3\%$). These results suggest that reasoning-capable models are better positioned to benefit from GRPO, learning not only to cite sources but also to assess answerability and provide more accurate and grounded responses overall.

Across all datasets, excessively high AR\% often correlates with lower refusal‐grounding (F1$_\text{GR}$) and citation quality (F1$_\text{GC}$), suggesting that blind answering harms grounded performance, while overly low AR\% (e.g.\ LLaMA‑3.1 Instruct) forfeits recall entirely. Qwen‑3 models under the Instruct setting exhibit near‐ceiling AR\% on ASQA, QAMPARI, and ELI5, but this comes at the expense of low F1$_\text{GR}$ and F1$_\text{GC}$, especially on QAMPARI (4.31\%/26.82\%) and ELI5 (36.22\%/18.54\%). In contrast, toggling to Reasoning prompts reduces AR\% by 10–20 points while boosting F1$_\text{GR}$ by 10–20 points and F1$_\text{GC}$ by 5–25 points across all datasets, yielding net TRUST gains of 5–10 points. Applying \NewMethod{} further amplifies these effects across nearly all models and settings (e.g., for Qwen‑3‑8b, F1$_\text{GR}$ on ASQA jumps from 41.97 to 66.67 and F1$_\text{GC}$ from 74.74 to 87.97), albeit sometimes at the cost of AR\%. One notable exception is Qwen‑3‑4b under the Instruct setting, where Ground‑GRPO actually lowers F1$_\text{AC}$ (62.50->51.25), F1$_\text{GR}$ (42.53->40.42) and overall TRUST (58.19→55.53), indicating that grounding in this case corrodes correctness rather than bolstering it on ASQA. LLaMA‑3.1 Instruct, which initially answers almost nothing (AR\% in the range of 1–4), flips to full coverage (100\% AR) under \NewMethod{} but sees only modest F1$_\text{GR}$/F1$_\text{GC}$ of 22–39\%, indicating that forced answering without reasoning degrades grounded fidelity. Finally, in the Reasoning setting, \NewMethod{} yields the most balanced trade‐off: AR\% sits in the 50–80 range, F1$_\text{AC}$ remains within 5 points of prompt‐only baselines, and F1$_\text{GR}$/F1$_\text{GC}$ improvements exceed 15 points in several cases, driving 10–15 point TRUST increases. These patterns demonstrate that combining reasoning with \NewMethod{} consistently optimizes grounded correctness and citation quality across diverse datasets (see \Cref{fig:reasoning-vs-instruct}).

\begin{AIbox}{TL;DR of RL Improves Groundedness}
\begin{itemize}[leftmargin=1em]
\item \NewMethod{} consistently improves overall groundedness (\metric{}) across all model types and QA datasets.
\item Reasoning models benefit more broadly, with gains across answer correctness, grounded refusal, and citation quality.
\item Improvements in grounding (\textbf{F1$_{\text{GC}}$}, \textbf{F1$_{\text{GR}}$}) sometimes come at the cost of answer correctness (\textbf{F1$_{\text{AC}}$}), especially for short-form QA or weaker base models.
\item Applying \NewMethod{} on reasoning models yields more balanced trade-offs and stronger overall alignment.
\end{itemize}
\end{AIbox}

\subsection{Decomposing Training Into Stages Improves Learning of Grounded Objectives}
Generating grounded responses involves three distinct subtasks: (1) determining whether a question is answerable given the evidence, (2) extracting the correct answer when it is, and (3) providing minimally sufficient citations to support the answer. When these objectives are trained jointly using GRPO, the resulting reward signal can be noisy, making it difficult for the model to optimize all three behaviors simultaneously.

To address this, we investigate a two-stage training strategy. In Stage 1, we train the model on a smaller dataset (300 samples) of answerable questions to focus on learning answer extraction and citation generation. In Stage 2, we introduce a larger, mixed dataset (3000 samples) containing both answerable and unanswerable questions, training the model to additionally learn to judge answerability.

\Cref{fig:stagewise-ablation} shows that this two-stage curriculum (Stage 1 + Stage 2) consistently improves groundedness compared to either stage alone. Specifically, it outperforms Stage 2-only training in 7 out of 9 model-dataset configurations. On average, two-stage training yields a 6.8\% gain over Stage 2-only and a 12.5\% gain over Stage 1-only training in \metric{}. The most significant improvement comes in \textbf{F1$_{\text{GR}}$} (Grounded Refusal), which increases by 15.8\% on average, suggesting that early training in grounded answer generation, followed by a broader training phase, helps models better integrate all aspects of groundedness.
\begin{figure*}[ht!]
    \centering
    \includegraphics[width=\textwidth]{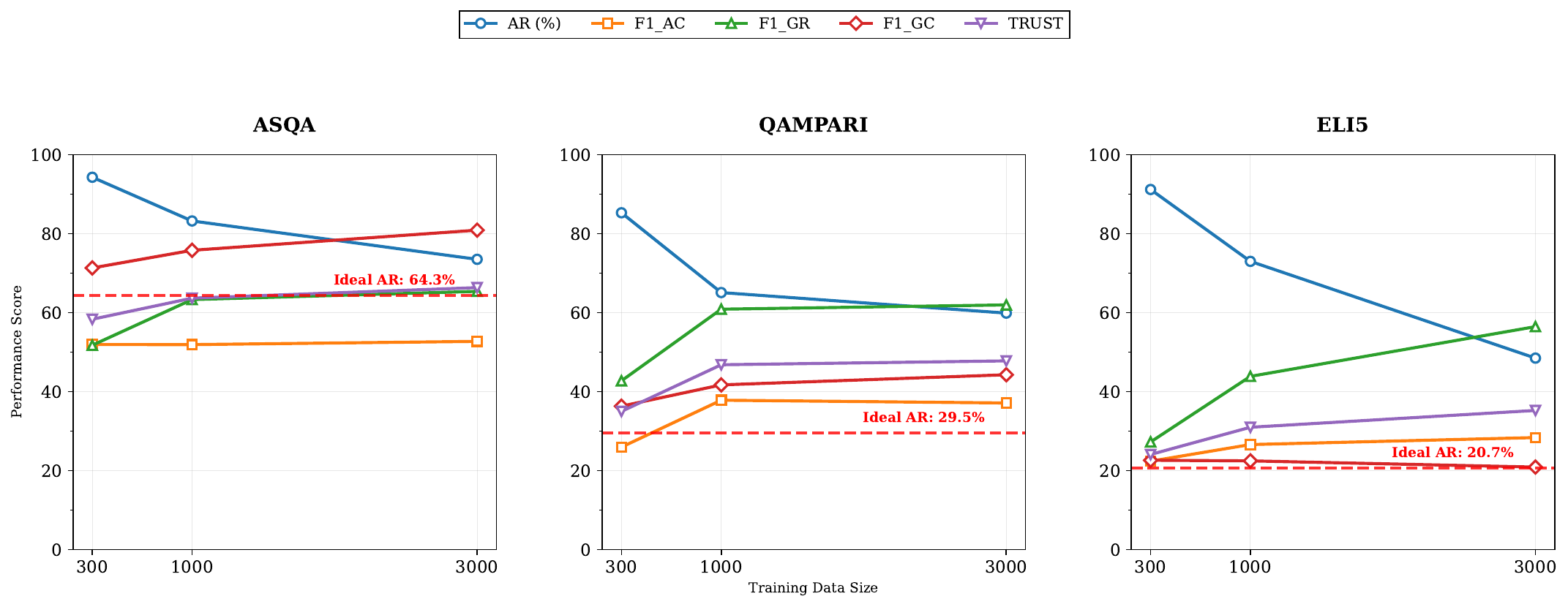}
    \caption{Performance trends across different training data sizes for three question-answering datasets. The analysis reveals contrasting patterns: while Answer Ratio (AR) consistently decreases with larger training sets across all datasets, most other metrics show improvement. F1$_\text{GR}$ (F1 score for grounded responses) demonstrates the most consistent positive correlation with data size, particularly pronounced in ASQA and QAMPARI datasets. F1$_\text{GC}$ (F1 score for grounded citations) shows steady improvement in ASQA and QAMPARI, but remains relatively stable in ELI5. F1$_\text{AC}$ (Answer Correctness) exhibits distinct behaviors: ASQA maintains stable performance (~52\%) regardless of training size, QAMPARI shows substantial improvement from 300 to 1,000 samples (25.96→37.84) before plateauing, and ELI5 demonstrates gradual but consistent gains (22.40→28.38). The TRUST metric exhibits moderate improvement with increased data size across all datasets. Notably, ELI5 displays the most volatile behavior, with AR showing the steepest decline and F1$_\text{GC}$ remaining nearly constant, suggesting dataset-specific characteristics may influence the effectiveness of scaling training data.}
    \label{fig:data_size_ablation}
\end{figure*}
\begin{AIbox}{TL;DR of Staged Training}
\begin{itemize}[leftmargin=1em]
\item Decomposing training into two stages yields more stable and targeted learning, helping models better optimize for grounded objectives.
\item Stage 1 focuses on grounded answer generation, while Stage 2 introduces answerability, resulting in stronger integration of all three behaviors.
\item This strategy especially improves \textbf{Grounded Refusal}, with a 15.8\% average gain, indicating better identification of unanswerable questions.
\end{itemize}
\end{AIbox}

\subsection{Distillation Helps}

Previous work in \method{} demonstrated that distilling high-quality responses from GPT-4 significantly improves the groundedness and citation quality of instruction-following models. They constructed a supervised fine-tuning (SFT) dataset with 10K samples and a Direct Preference Optimization (DPO) dataset with approximately 17K samples, both based on GPT-4 outputs. Models trained on this distilled data using SFT and DPO showed substantial gains in response trustworthiness. Building on this, we investigate how reinforcement learning via on-policy GRPO compares to distillation-based approaches, and whether distillation before or after GRPO further improves model performance. Given that the Trust-Align Dataset was built for instruct models, we only investigate the impact of distillation on instruct variants of models.

\begin{figure*}[ht]
\centering
\includegraphics[width=\textwidth]{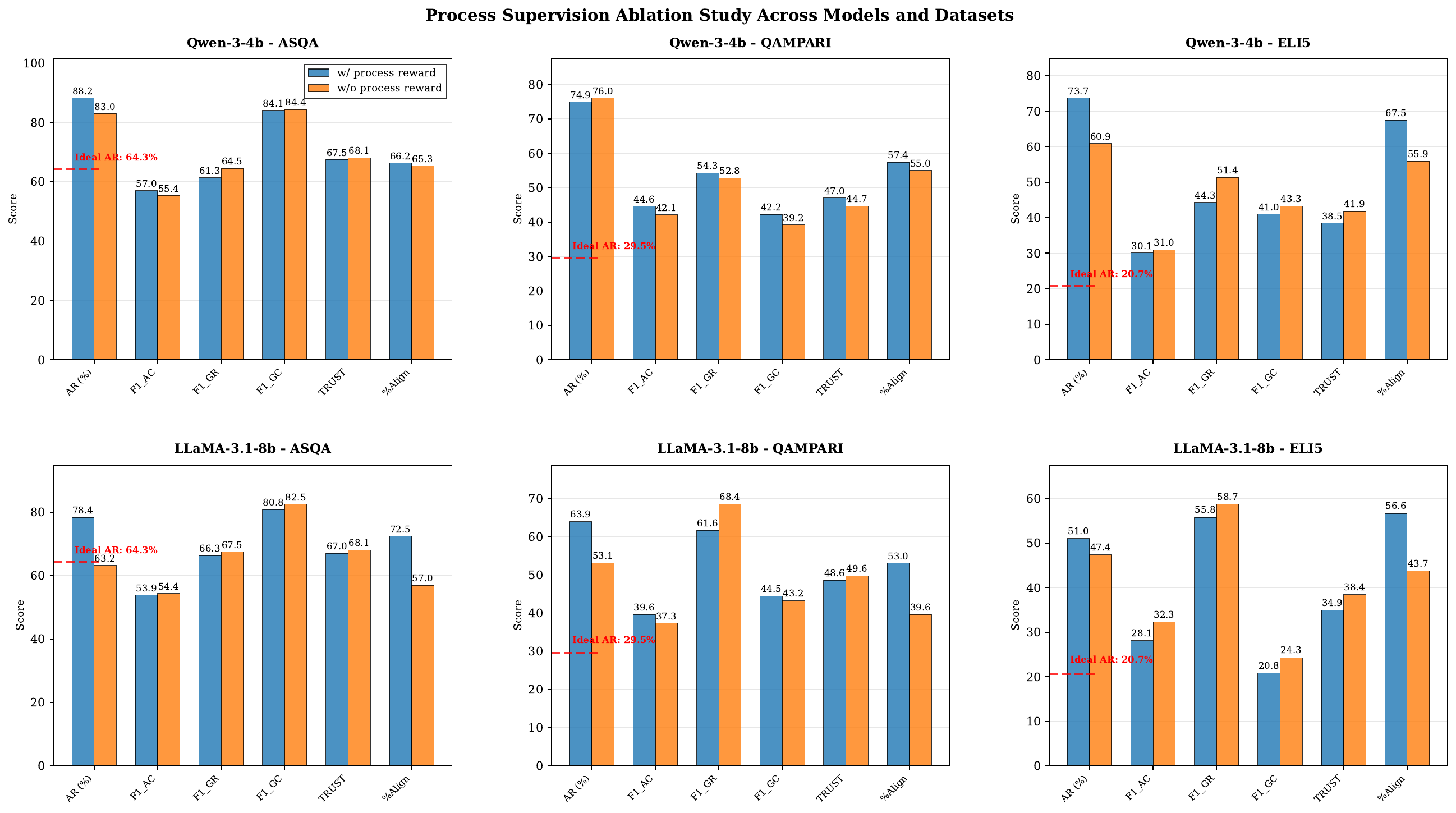}
\caption{Process supervision ablation study across two language models and three question-answering datasets. The comparison between models with and without process reward supervision reveals dataset-dependent effects. Notably, F1$_\text{GR}$ (grounded responses) demonstrates contrasting behaviors: while Qwen-3-4b shows slight decreases with process rewards in ASQA and QAMPARI, LLaMA-3.1-8b exhibits significant improvements in QAMPARI (61.63 vs 68.44) but decreases in other datasets. The \%Align metric, indicating alignment between reasoning steps and final answers, generally favors process reward supervision, particularly for Qwen-3-4b across all datasets and LLaMA-3.1-8b in ASQA and ELI5. These results suggest that process supervision benefits vary significantly by model architecture and dataset complexity, \textbf{with reasoning alignment showing the most consistent improvements}.}
\label{fig:process_supervision_ablation}
\end{figure*}

\subsubsection{RL vs. Distillation}
\Cref{table:Trust-Align} shows that instruct models trained with the Trust-Align SFT and DPO pipeline outperform those trained solely with \NewMethod{}. This performance gap may stem from several factors. First, instruct models appear less capable of effectively leveraging on-policy learning compared to reasoning models. Second, the Trust-Align approach benefits from a substantially larger training corpus—over 27K samples—while our GRPO experiments use only 3K examples. Finally, the high-quality supervision signal provided by GPT-4 through distillation may be especially valuable for instruction-following models, particularly when their initial performance is insufficient to reliably obtain reward signals during RL training. These observations highlight the complementary strengths of distillation and GRPO for instruction-tuned models, motivating us to investigate whether applying GRPO on top of distillation can further enhance groundedness and citation quality.

\subsubsection{Applying \NewMethod{} to Distilled Models}
As shown in \Cref{table:Trust-Align}, combining distillation with \NewMethod{} leads to notable performance improvements across all three QA benchmarks. When comparing distilled models trained with \NewMethod{} (SFT $\to$ \NewMethod{}) against models trained with \NewMethod{} alone, we observe significant gains in \metric{}: a 7.5\% improvement on ASQA, 23.0\% on QAMPARI, and 14.8\% on ELI5. These results suggest that distillation serves as a strong initialization for reinforcement learning, improving the model’s ability to earn rewards and enabling more effective alignment toward grounded response generation.

However, applying \NewMethod{} to already distilled models (SFT and SFT $\to$ DPO) shows mixed results when compared to the distilled models. On long-form QA tasks such as ELI5, which demand detailed, explanatory answers grounded in retrieved documents, \NewMethod{} improves performance by 4.2\% on average. Specifically, SFT models improve by 4.4\%, and SFT $\to$ DPO models by 4.2\%. In contrast, on short-form QA tasks like QAMPARI, which require concise, list-style answers with exact formatting, performance declines significantly: SFT models drop by 9.7\%, and SFT $\to$ DPO models by 7.2\%.

A closer look reveals that the most significant degradation occurs in the Answer Correctness component of \metric{}, with an average drop of 6.2\%, and up to 10\% on QAMPARI. On ASQA, We also observe an average of 6.6\% from the SFT models and a 9.9\% drop from the SFT $\to$ DPO models when they are further trained with \NewMethod{}. This decline appears to stem from two main factors. First, both ASQA and QAMPARI benchmarks rely on exact string matching to assess answer correctness, making them particularly sensitive to minor formatting differences such as extra punctuation or slight variations in phrasing. While the model's responses may be factually correct, such surface-level mismatches can lead to lower scores. Second, the reward function used in \NewMethod{} is relatively strict, which may cause the model to adopt a more conservative answering style returning only one or two confident answers even when the documents contain more, thereby reducing the Answer Correctness score.

Importantly, since ASQA and QAMPARI feature relatively short, factoid-style answers, the benefits of \NewMethod{} appear limited when applied on top of already well-performing distilled models. In these settings, the task of extracting concise factual answers is relatively straightforward, and distillation alone often suffices. As a result, the additional optimization from \NewMethod{} can introduce unnecessary cautiousness or formatting inconsistencies, ultimately hurting performance. In contrast, long-form tasks like ELI5 demand more complex reasoning, nuanced answer synthesis, and careful grounding, which are areas where the alignment signal from \NewMethod{} is more impactful. These findings suggest that while GRPO complements distillation in open-ended, generative tasks, it may offer diminishing returns, or even degrade performance, on structured, short-form QA tasks where distillation is already highly effective.

Nevertheless, we observe an average gain of 2.4\% in \textbf{F1$_{\text{GC}}$} across models and datasets, indicating that \NewMethod{} improves the model’s ability to provide accurate, non-redundant citations. This suggests that even in cases where answer correctness plateaus or slightly declines, \NewMethod{} can still enhance the faithfulness and clarity of citation attribution.

Although applying \NewMethod{} to distilled models generally leads to improved performance on ELI5, LLaMA-3.1-8b is a notable exception. The reason for this remains unclear. One possible explanation is that LLaMA-3.1-8b has undergone a different post-training process compared to Qwen models, which were fine-tuned using reasoning-focused, RL-based methods. A similar trend is observed with Gemma2-9b, where the performance gain on ELI5 is minimal. We also see this pattern in another longform QA dataset, ExpertQA, as discussed in \Cref{sec:ood}.

Finally, we also compare the ordering of distillation and \NewMethod{} training. Across all models and datasets, we find that applying distillation before \NewMethod{} ( SFT $\to$ \NewMethod{} and SFT $\to$ DPO $\to$ \NewMethod{}) consistently outperforms the reverse order (\NewMethod{} $\to$ SFT and \NewMethod{} $\to$ SFT $\to$ DPO) by a margin of 15\%. This result aligns with our earlier observations: distillation provides a strong, high-quality initialization that enables the model to better leverage the reward signal during reinforcement learning. In contrast, when \NewMethod{} is applied first, the model may not yet possess the necessary grounding or fluency to benefit fully from subsequent supervised fine-tuning, leading to suboptimal alignment. These findings reinforce the importance of sequencing in training pipelines and highlights that distillation, when used as a precursor to reinforcement learning, provides a more effective foundation for improving trustworthiness in instruct models.

\begin{AIbox}{TL;DR of \NewMethod{} with Distilled Models}
\begin{itemize}[leftmargin=1em]
\item \NewMethod{} complements distillation on longform QA datasets such as ELI5 and ExpertQA.
\item On shortform QA datasets like ASQA and QAMPARI, \NewMethod{} offers no noticeable benefits when applied to distilled models.
\item The effectiveness of \NewMethod{} with distilled models depends on the underlying backbone LLM. It performs better with Qwen-3 models compared to LLaMA-3.1 and Gemma-2. This may be attributed to Qwen-3 models receiving reasoning-focused RL-based post-training, unlike the latter two.
\item Ordering of distillation and \NewMethod{} matters. Applying Distillation Before \NewMethod{} shows a significant performance increase compared to applying \NewMethod{} first.
\end{itemize}
\end{AIbox}

\subsection{Out-of-Distribution (OOD) Performance}
\label{sec:ood}
\Cref{table: expertqa_general} presents results on the ExpertQA benchmark, evaluating the generalization ability of various model configurations. Across all three models' reasoning variants, we observe that \NewMethod{} with two-stage training (Stage 1 + Stage 2) consistently achieves higher TRUST scores compared to models trained using only one stage, highlighting the benefits of progressive alignment.

Moreover, reinforcement learning via \NewMethod{} consistently enhances ExpertQA performance in 5 out of 6 instruct model configurations. Notably, applying \NewMethod{} on top of SFT or SFT $\to$ DPO leads to substantial gains in F1$_{\text{GC}}$, up to 17\% improvement, showing that \NewMethod{} helps models better support their claims with grounded evidence even in unfamiliar domains. These improvements also translate to overall TRUST score gains, such as a 10-point increase in the Qwen-3-4B instruct variant when \NewMethod{} is applied on top of SFT $\to$ DPO. Given that ExpertQA is a longform dataset, this finding further supports our point that applying \NewMethod{} over distilled models would greatly improve performance in complex open-ended tasks.

In contrast, standard instruct and reasoning models without alignment perform poorly on ExpertQA, despite high answer rates (AR), suggesting overconfidence and a lack of reliable refusal behavior. In contrast, the improved performance of \NewMethod{}-trained and two-stage models highlights their ability to more accurately recognize when not to answer and to cite appropriately, a key requirement for trustworthy generalization in OOD settings.

\begin{table}[t]
\caption{Results on ExpertQA. \NewMethod{} consistency improves the performance when applied to base or distilled model variants.}
\label{table: expertqa_general}

\centering
\resizebox{\linewidth}{!}{
\begin{tabular}{ll*{5}{c}}
\toprule
\textbf{Model} & \textbf{Type} & \textbf{AR (\%)} & \textbf{F1$_{\text{AC}}$} & \textbf{F1$_{\text{GR}}$} & \textbf{F1$_{\text{GC}}$} & \textbf{TRUST} \\ 

\midrule

\multirow{11}{*}{\mystack{Qwen-3}{-4b}}
& \textit{Instruct Variant} \\
& Prompt (Instruct) & 98.39 & 35.31 & 26.32 & 52.80 & 38.14 \\
& \NewMethod{} (Instruct) & 98.34 & 32.94 & 26.29 & 46.27 & 35.17 \\
& \cellcolor{rowgray}SFT & \cellcolor{rowgray}35.04 & \cellcolor{rowgray}20.48 & \cellcolor{rowgray}63.95 & \cellcolor{rowgray}42.84 & \cellcolor{rowgray}42.43 \\
& \cellcolor{rowgray}\(\text{SFT} \to \NewMethod{}\) & \cellcolor{rowgray}27.39 & \cellcolor{rowgray}20.56 & \cellcolor{rowgray}67.14 & \cellcolor{rowgray}59.37 & \cellcolor{rowgray}49.02 \\
 \cdashlinelr{2-7}
& \cellcolor{rowpeach}\(\text{SFT} \to \text{DPO}\) & \cellcolor{rowpeach}28.08 & \cellcolor{rowpeach}22.33 & \cellcolor{rowpeach}63.55 & \cellcolor{rowpeach}50.27 & \cellcolor{rowpeach}45.39 \\
& \cellcolor{rowpeach}\(\text{SFT} \to \text{DPO} \to \NewMethod{}\) & \cellcolor{rowpeach}16.78 & \cellcolor{rowpeach}24.67 & \cellcolor{rowpeach}67.62 & \cellcolor{rowpeach}75.62 & \cellcolor{rowpeach}55.97 \\
\cdashlinelr{2-7}
& \textit{Reasoning Variant} \\
& Prompt (Reasoning) & 82.47 & 39.23 & 42.23 & 40.77 & 40.74 \\
& \NewMethod{} w/ Stage 1 only & 94.26 & 35.41 & 31.46 & 49.98 & 38.95 \\
& \NewMethod{} w/ Stage 2 only & 85.07 & 35.69 & 41.68 & 52.37 & 43.24 \\
& \NewMethod{} w/ Stage 1 + 2 & 68.51 & 33.90 & 52.86 & 52.29 & 46.35 \\

\midrule

\multirow{11}{*}{\mystack{Qwen-3}{-8b}}
& \textit{Instruct Variant} \\
& Prompt (Instruct) & 98.62 & 37.01 & 25.95 & 54.19 & 39.05 \\
& \NewMethod{} (Instruct) & 29.14 & 29.58 & 70.63 & 82.54 & 60.92 \\  
& \cellcolor{rowgray}SFT & \cellcolor{rowgray}19.59 & \cellcolor{rowgray}33.03 & \cellcolor{rowgray}68.41 & \cellcolor{rowgray}58.85 & \cellcolor{rowgray}53.43 \\
& \cellcolor{rowgray}\(\text{SFT} \to \NewMethod{}\) & \cellcolor{rowgray}35.27 & \cellcolor{rowgray}29.05 & \cellcolor{rowgray}71.43 & \cellcolor{rowgray}66.95 & \cellcolor{rowgray}55.81 \\
 \cdashlinelr{2-7}
& \cellcolor{rowpeach}\(\text{SFT} \to \text{DPO}\) & \cellcolor{rowpeach}34.30 & \cellcolor{rowpeach}29.50 & \cellcolor{rowpeach}70.12 & \cellcolor{rowpeach}60.24 & \cellcolor{rowpeach}53.29 \\
& \cellcolor{rowpeach}\(\text{SFT} \to \text{DPO} \to \NewMethod{}\) & \cellcolor{rowpeach}20.01 & \cellcolor{rowpeach}26.08 & \cellcolor{rowpeach}68.87 & \cellcolor{rowpeach}77.26 & \cellcolor{rowpeach}57.40 \\
\cdashlinelr{2-7}
& \textit{Reasoning Variant} \\
& Prompt (Reasoning) & 82.63 & 40.29 & 43.67 & 42.72 & 42.23 \\
& \NewMethod{} w/ Stage 1 only & 99.42 & 36.74 & 25.94 & 50.83 & 37.83 \\
& \NewMethod{} w/ Stage 2 only & 80.82 & 38.88 & 45.56 & 48.95 & 44.46 \\
& \NewMethod{} w/ Stage 1 + 2 & 52.70 & 39.64 & 60.92 & 57.09 & 52.55 \\

\midrule

\multirow{11}{*}{\mystack{LLaMA-3.1}{-8b}}
& \textit{Instruct Variant} \\
& Prompt (Instruct) & 90.36 & 36.24 & 34.26 & 48.16 & 39.55 \\
& \NewMethod{} (Instruct) & 100.00 & 19.26 & 23.92 & 47.51 & 30.23 \\  
& \cellcolor{rowgray}SFT & \cellcolor{rowgray}23.93 & \cellcolor{rowgray}26.84 & \cellcolor{rowgray}69.89 & \cellcolor{rowgray}62.00 & \cellcolor{rowgray}52.91 \\
& \cellcolor{rowgray}\(\text{SFT} \to \NewMethod{}\) & \cellcolor{rowgray}23.84 & \cellcolor{rowgray}24.10 & \cellcolor{rowgray}70.32 & \cellcolor{rowgray}58.76 & \cellcolor{rowgray}51.06 \\
 \cdashlinelr{2-7}
& \cellcolor{rowpeach}\(\text{SFT} \to \text{DPO}\) & \cellcolor{rowpeach}7.01 & \cellcolor{rowpeach}17.27 &\cellcolor{rowpeach} 57.25 & \cellcolor{rowpeach}75.05 & \cellcolor{rowpeach}49.86 \\
& \cellcolor{rowpeach}\(\text{SFT} \to \text{DPO} \to \NewMethod{}\) & \cellcolor{rowpeach}11.85 & \cellcolor{rowpeach}18.85 & \cellcolor{rowpeach}63.91 & \cellcolor{rowpeach}71.08 & \cellcolor{rowpeach}51.28\\
\cdashlinelr{2-7}
& \textit{Reasoning Variant} \\
& Prompt (Reasoning) & 97.97 & 33.02 & 26.59 & 23.48 & 27.70 \\
& \NewMethod{} w/ Stage 1 only & 98.48 & 32.07 & 25.98 & 28.50 & 28.85 \\
& \NewMethod{} w/ Stage 2 only & 97.69 & 31.80 & 27.01 & 32.96 & 30.59 \\
& \NewMethod{} w/ Stage 1 + 2 & 45.46 & 37.99 & 63.77 & 39.44 & 47.06 \\

\bottomrule
\end{tabular}
}
\end{table}

\subsection{Data Size Ablation}

We investigated the impact of Stage 2 dataset size in \NewMethod{} on model performance with 3 different dataset sizes: 300, 1000 and 3000 samples. As shown in \Cref{fig:data_size_ablation}, increasing the number of training samples in Stage 2 resulted in a consistent improvement in overall \metric{} and across each of its individual components. These results indicate that larger datasets provide a stronger supervision signal for aligning models with grounded behavior and that \NewMethod{} can scale effectively with additional training data.

\begin{figure*}[ht!]
\centering
\includegraphics[width=\textwidth]{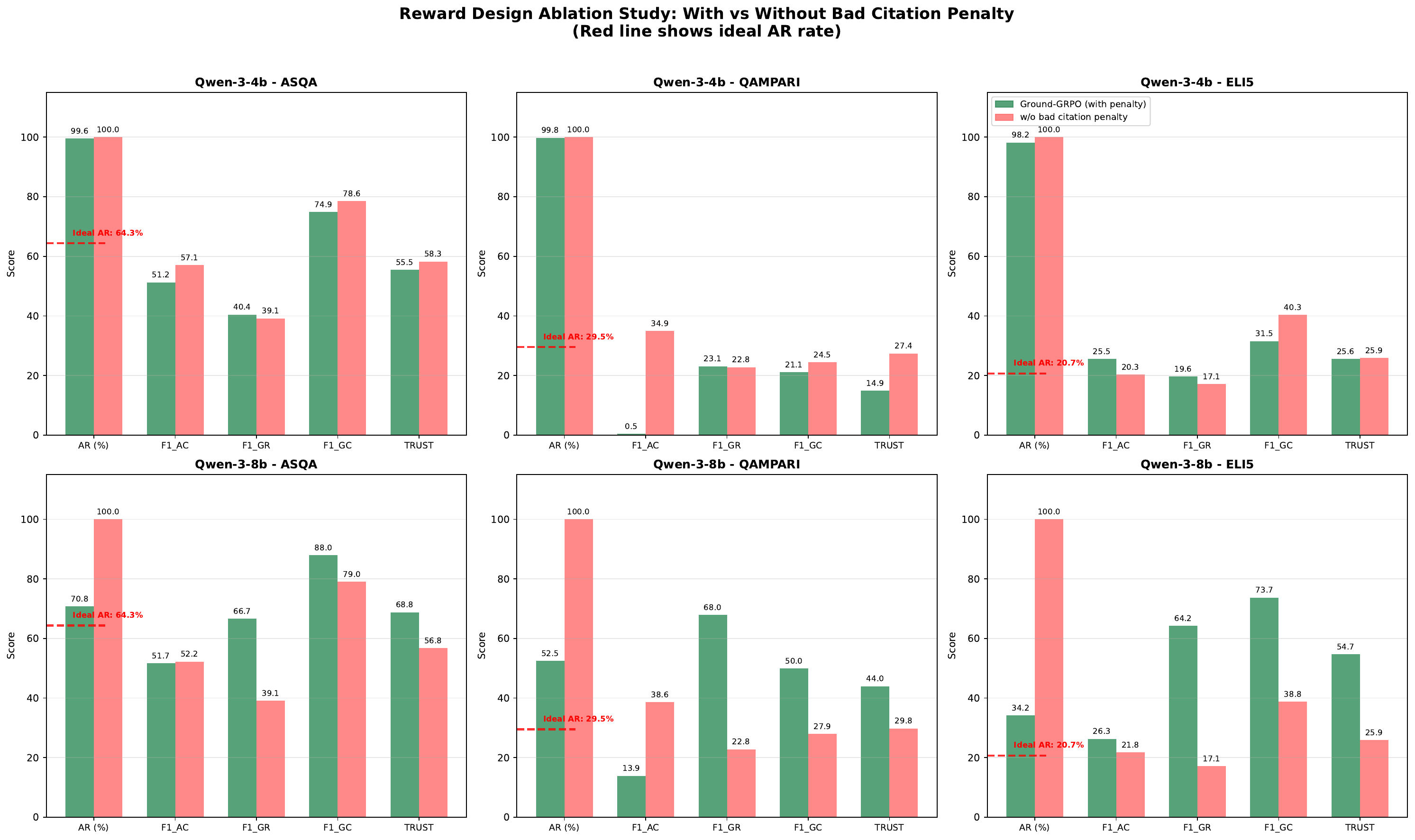}
\caption{
\textbf{Reward Design Ablation Study: Impact of Bad Citation Penalty on Model Performance.} 
We compare Ground-GRPO training with and without the bad citation penalty component across two model variants (Qwen-3-4b and Qwen-3-8b) on three datasets. The red dashed lines indicate the ideal Answer Rate (AR) for each dataset, computed as the ratio of answerable to total questions: ASQA (64.4\%), QAMPARI (29.5\%), and ELI5 (20.7\%). 
\textbf{Key findings:} (1) Removing the bad citation penalty consistently drives AR to near 100\%, significantly overshooting the ideal rates and indicating over-answering behavior across both models. (2) The impact on other metrics shows \textit{contrasting patterns between models}: For Qwen-3-4b, removing the penalty generally improves F1$_{\text{AC}}$ and TRUST scores with minimal impact on F1$_{\text{GR}}$, while for Qwen-3-8b, it dramatically degrades F1$_{\text{GR}}$ performance (e.g., from 67.96 to 22.78 on QAMPARI) despite improving F1$_{\text{AC}}$. (3) These \textit{model-dependent effects} suggest that optimal reward configuration varies by model architecture and capacity, with larger models (8b) being more sensitive to citation penalty removal. (4) The penalty serves different roles across models: acting as a mild regularizer for smaller models but as a crucial constraint preventing performance collapse in larger models. This highlights the importance of model-specific reward tuning in retrieval-augmented generation training.
}
\label{fig:reward-ablation}
\end{figure*}
\subsection{Process Supervision Analysis}
\label{sec:process-supervision-analysis}

In some experiments, we observed that the model’s internal reasoning was occasionally misaligned with its final answer. In particular, there were instances where the reasoning chain indicated that the model had found sufficient evidence to answer the question, but the final response incorrectly refused to answer, claiming that no relevant information was found  (see \Cref{sec:process_misalignment_example} for an illustrative example). This mismatch suggests that the model’s decision-making during generation is not always consistent across the reasoning and final answer, highlighting the potential need for more explicit process supervision. 

To this end, we introduce a process reward using an NLI model that evaluates whether the model’s reasoning supports its final decision to answer or refuse:

\begin{align}
R_{\text{process}} = \texttt{nli\_score}
\end{align}

Here, \texttt{nli\_score} reflects the degree to which the reasoning trace entails the final decision (either answering or refusing).

To measure the alignment between reasoning and final output, we define a simple consistency metric:

\begin{align}
\text{\%Align} = \frac{1}{N} \sum_{i=1}^{N} \mathbf{1}\left[\text{NLI}(r_i, a_i) \rightarrow \text{entailment}\right]
\end{align}

\Cref{fig:process_supervision_ablation} shows that incorporating this process reward results in a modest decrease in \metric{} in 5 out of 6 configurations, with an average drop of 1.2\%. However, we observe a notable average improvement of 8\% in \%Align, indicating that the NLI-based process reward improves the consistency between reasoning and final answers. These findings suggest that while process supervision introduces trade-offs in overall groundedness, it holds promise for reducing reasoning-answer misalignment. Future work could focus on refining this approach to maintain or enhance groundedness while promoting internal consistency.

\begin{AIbox}{TL;DR of Process Supervision}
\begin{itemize}[leftmargin=1em]
\item We discovered that RL-trained models sometimes generate reasoning that contradicts the final answer, indicating internal misalignment.
\item We introduce a process reward based on NLI to encourage consistency between reasoning and final output.
\item This improves reasoning-answer alignment by 8\% on average, but leads to a small (1.2\%) drop in overall \metric{}.
\end{itemize}
\end{AIbox}

\subsection{Reward Design Ablation}

We conducted an ablation study to examine the reward design used in \NewMethod{} training, focusing on the impact of the bad citation penalty. As shown in \Cref{fig:reward-ablation}, removing this penalty has a substantial influence on model behavior. First, we observe that eliminating the bad citation penalty consistently drives the AR close to 100\%, indicating a strong tendency toward over-answering, even when the input documents do not support a grounded response. Second, the effect on other evaluation metrics varies significantly by model size. For the smaller Qwen-3-4B model, removing the penalty slightly improves \textbf{F1$_{\text{AC}}$} and overall \metric{} scores, with minimal effect on \textbf{F1$_{\text{GC}}$}. In contrast, for the larger Qwen-3-8B model, removing the citation penalty leads to a drastic drop in \textbf{F1$_{\text{GC}}$} (e.g., from 67.96 to 22.78 on QAMPARI), despite modest gains in \textbf{F1$_{\text{GC}}$}. These findings suggest that the role of the bad citation penalty is model-dependent: it acts as a light regularizer in smaller models, but as a critical constraint in larger models, where its absence can lead to citation failure and collapse in groundedness. Overall, this highlights the importance of model-specific reward tuning for grounded response generation.

\section{Related works}

\subsection{Attributable Retrieval-Augmented Generation}
Retrieval-Augmented Generation (RAG) has become a foundational approach for enhancing factual accuracy by grounding LLM outputs in retrieved evidence \cite{karpukhin-etal-2020-dense, lewis2021retrievalaugmented, gao2023retrieval}. However, models can still be misled by irrelevant passages, leading to hallucinations and ungrounded outputs \cite{pmlr-v202-shi23a, yoran2024makingretrievalaugmentedlanguagemodels, xu2023recompimprovingretrievalaugmentedlms}.

\subsection{Training-Based Grounded Generation}
Approaches to enhancing grounding through training can be categorized into training-free and training-based paradigms. Training-free methods focus on citation reflexivity, such as in-context learning with citation examples \cite{gao2023enabling}, post-hoc retrieval for attribution \cite{gao-etal-2023-rarr, li2024citationenhanced}, or even chain-of-thought prompting to improve citation alignment \cite{Ji_Liu_Du_Ng_2024}. Training-based methods include supervised fine-tuning (SFT) on citation-rich data \cite{asai2024selfrag, slobodkin2024attribute, xia2024groundsentenceimprovingretrievalaugmented, ye2024effective} and preference-based learning via RLHF and Direct Preference Optimization (DPO) \cite{ouyang2022traininglanguagemodelsfollow, rafailov2024directpreferenceoptimizationlanguage}. \citeauthor{huang2024training} use fine-grained rewards and PPO to refine attribution, and \citeauthor{li2024improving} develop a modified DPO framework to enhance citation control. Unlike these, our method \method{} introduces a staged GRPO training that decomposes grounding behaviors—first optimizing answer and citation quality, then refusal, without requiring gold reasoning traces.

\section{Recommendations}
Based on our preliminary testing with a few LLMs, we provide the following recommendations to enhance the accuracy and grounding of RAG systems. We strongly encourage readers to rigorously validate these recommendations across different LLM families and model sizes.

\begin{AIbox}{Recommendations}
\begin{itemize}[leftmargin=1em]
\item Use reasoning models for RAG, as they improve both answer generation accuracy and grounding.
\item Design relevant reward functions and apply GRPO for on-policy enhancement of LLM accuracy and grounding in RAG. Since not all reward functions benefit every model, perform model-specific evaluations.
\item Knowledge distillation from larger models such as GPT-4 can boost the accuracy and grounding of RAG systems.
\item Apply GRPO to distilled models for long-form question answering. For short-form or simpler questions, we do not recommend applying GRPO to distilled models, as our experiments show no improvement and, in some cases, a decline in performance.
\end{itemize}
\end{AIbox}

\bibliography{aaai2026}

\appendix

\onecolumn

\section{Metric Details} \label{sec:metrics_app}

We report performance using \metric{}, a composite metric that evaluates the trustworthiness of model responses across three dimensions: response truthfulness, factual accuracy, and attribution groundedness.

\subsection{Response Truthfulness}

This component assesses whether the model correctly answers questions when evidence is available and appropriately refuses when it is not. It is computed as a macro-average of two F1 scores:

\paragraph{Grounded Refusal [\textbf{F1$_{\text{GR}}$}]:}
We define two subsets: $A_g$ and $\neg A_g$ as the ground truth sets of answerable and unanswerable questions, and $A_r$ and $\neg A_r$ as the sets of questions the model answers and refuses, respectively.

\begin{itemize}
\item \textbf{F1\textsubscript{ref}}: Measures the model's ability to correctly refuse unanswerable questions.
\item \textbf{F1\textsubscript{ans}}: Measures the model's ability to correctly answer answerable questions.
\end{itemize}

These are computed via precision and recall and combined into:
\begin{align}
\textbf{F1$_{\text{GR}}$} &= \frac{1}{2}(\text{F1}_{\text{ref}} + \text{F1}\_{\text{ans}})
\end{align}

This score captures the balance between under-refusal and over-refusal behavior.

\subsection{Factual Accuracy}

This component evaluates whether the model's answers are factually accurate and grounded in the provided evidence.

\paragraph{Answer Correctness [\textbf{F1$_{\text{AC}}$}]:}
For a question $q$, let $A_G$ be the set of gold claims, $A_D$ the claims supported by retrieved documents, and $A_R$ the claims generated in the model's response. The calibrated answer correctness for $q$ is:

\begin{align}
\text{AC}^{q} = \frac{|A_G \cap A_D \cap A_R|}{|A_G \cap A_D|}
\end{align}

This restricts evaluation to claims verifiable from retrieved evidence. To aggregate across the dataset, we compute precision-oriented (\textbf{P\textsubscript{AC}}) and recall-oriented (\textbf{R\textsubscript{AC}}) variants, depending on whether the denominator is the number of answered or answerable questions. The combined F1 is:

\begin{align}
\textbf{F1$_{\text{AC}}$} = \frac{2 \text{ } \text{P}_{\text{AC}} \cdot \text{R}_{\text{AC}}}{\text{P}_{\text{AC}} + \text{R}_{\text{AC}}}
\end{align}

This metric discourages hallucination by rewarding answers that are both accurate and present in the evidence.

\subsection{Attribution Groundedness}

This component measures whether cited documents genuinely support the statements they are associated with.

\paragraph{Grounded Citation F1 [\textbf{F1$_{\text{GC}}$}]:}
Each response consists of statements $\mathcal{S}$ and citations $\mathcal{C}$. We assess:
\begin{itemize}
    \item \textbf{Recall}: Whether the full set of cited documents supports each statement.
    \item \textbf{Precision}: Whether each individual citation contributes uniquely to the support of a statement.
\end{itemize}

We compute:

\begin{align}
\textbf{F1$_{\text{GC}}$} = \frac{2 \cdot \text{P}_{\text{cite}} \cdot \text{R}_{\text{cite}}}{\text{P}_{\text{cite}} + \text{R}_{\text{cite}}}
\end{align}

This score penalizes both spurious citations and incomplete attribution.

\subsection{Overall Metric}

We combine the three components into a single aggregate trustworthiness score:

\begin{align}
\metric{} = \frac{1}{3} (\textbf{F1$_{\text{GR}}$} + \textbf{F1$_{\text{AC}}$} + \textbf{F1$_{\text{GC}}$})
\end{align}

This composite score enables consistent model ranking across correctness, calibration, and citation quality.

\section{Need for Process Supervision}
\label{sec:process_misalignment_example}
In our analysis, we observed cases where the model's final answer was a refusal, despite the reasoning trace indicating a clear and confident answer. This mismatch between the model’s internal reasoning and final output suggests that the model may not be using its reasoning trace to inform its final decision, highlighting a gap in alignment.

For instance, consider the following example:

\begin{description}
  \item[Q:] Who has the highest goals in world football?
  \item[Reasoning Trace:] ... Since Bican's is the official count, he's the highest. Chitalu's goals might not be official, so he's not the answer. So, the answer is Josef Bican with 805 goals, cited by RSSF and IFFHS.
  \item[Final Answer:] I'm sorry, but I couldn't find an answer to your question in the search results.
\end{description}

Despite clearly identifying the correct answer in the reasoning trace, the model ultimately refused to answer. This inconsistency may stem from a lack of explicit supervision over how the reasoning trace should support the final decision. In such cases, the model might learn to produce refusals in the final output without developing the proper reasoning chain to justify that refusal.

To address this issue, we introduced a process-level reward based on Natural Language Inference (NLI) to better align the model’s reasoning trace with its final decision (see Section~\ref{sec:process-supervision-analysis}). The example above illustrates the motivation for this process reward: encouraging the model to maintain consistency between its intermediate reasoning and its final answer.

\section{\method{}}
\begin{figure*}[h]
    \centering
    \includegraphics[width=\textwidth]{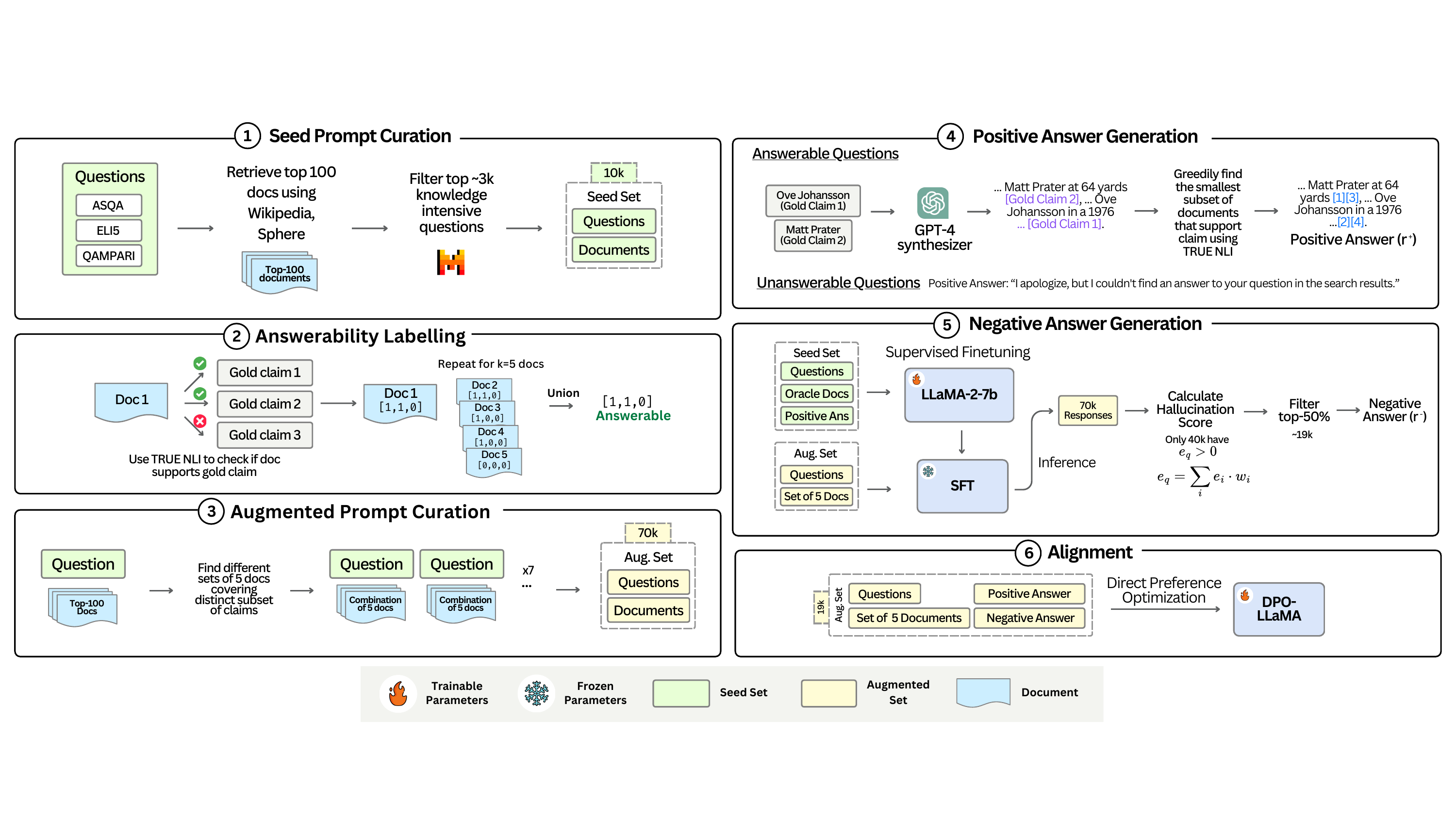}
    \caption{The pipeline of \method{}.}
    \label{fig:trust-align}
\end{figure*}
\end{document}